\documentclass[runningheads]{llncs}
\usepackage{graphicx}
\usepackage{amsmath,amssymb} 
\usepackage{color}
\usepackage{algpseudocode,algorithmicx,algorithm}
\usepackage[width=122mm,left=12mm,paperwidth=146mm,height=193mm,top=12mm,paperheight=217mm]{geometry}
\begin{document}
\pagestyle{headings}
\mainmatter

\newcommand\blfootnote[1]{%
  \begingroup
  \renewcommand\thefootnote{}\footnote{#1}%
  \addtocounter{footnote}{-1}%
  \endgroup
}

\renewcommand{\thefootnote}{\fnsymbol{footnote}}

\title{Deep learning based fence segmentation and removal from an image using a video sequence} 



\author{Sankaraganesh Jonna$^1$, Krishna K. Nakka$^{2,\bf{\ast}}$, and Rajiv R. Sahay$^2$}
\institute{$^1$Department of Computer Science and Engineering,\\
	$^2$Department of Electrical Engineering, \\
	$^{1,2}$Indian Institute of Technology Kharagpur, India\\
		\email{ \{sankar9.iitkgp, krishkanth.92, sahayiitm\}@gmail.com}
		}

\maketitle

\footnote[1]{The second author contributed while pursing masters at IIT Kharagpur}
\begin{abstract}
	Conventiona approaches to image de-fencing use multiple adjacent frames for segmentation of fences in the reference image and are limited to restoring images of static scenes only. In this paper, we propose a de-fencing algorithm for images of dynamic scenes using an occlusion-aware optical flow method. We divide the problem of image de-fencing into the tasks of automated fence segmentation from a single image, motion estimation under known occlusions and fusion of data from multiple frames of a captured video of the scene. Specifically, we use a pre-trained convolutional neural network to segment fence pixels from a single image. The knowledge of spatial locations of fences is used to subsequently estimate optical flow in the occluded frames of the video for the final data fusion step. We cast the fence removal problem in an optimization framework by modeling the formation of the degraded observations. The inverse problem is solved using fast iterative shrinkage thresholding algorithm (FISTA). Experimental results show the effectiveness of proposed algorithm.
\end{abstract}

\section{Introduction}
\label{sec:introduction}
\begin{figure*}[t]
	\centering
	\begin{tabular}{c c c c}
		\includegraphics[width=2.9cm]{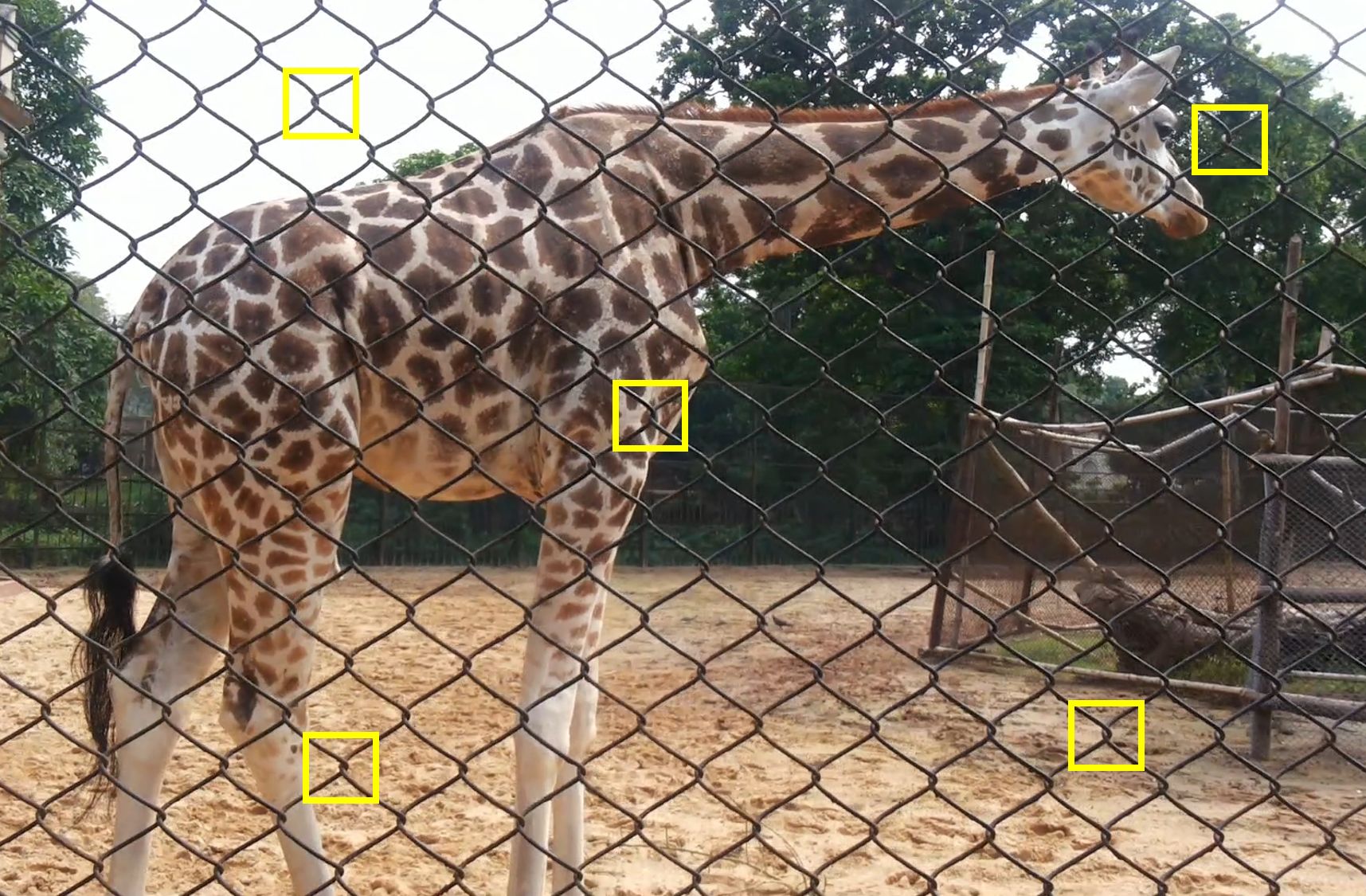}&
		\includegraphics[width=2.9cm]{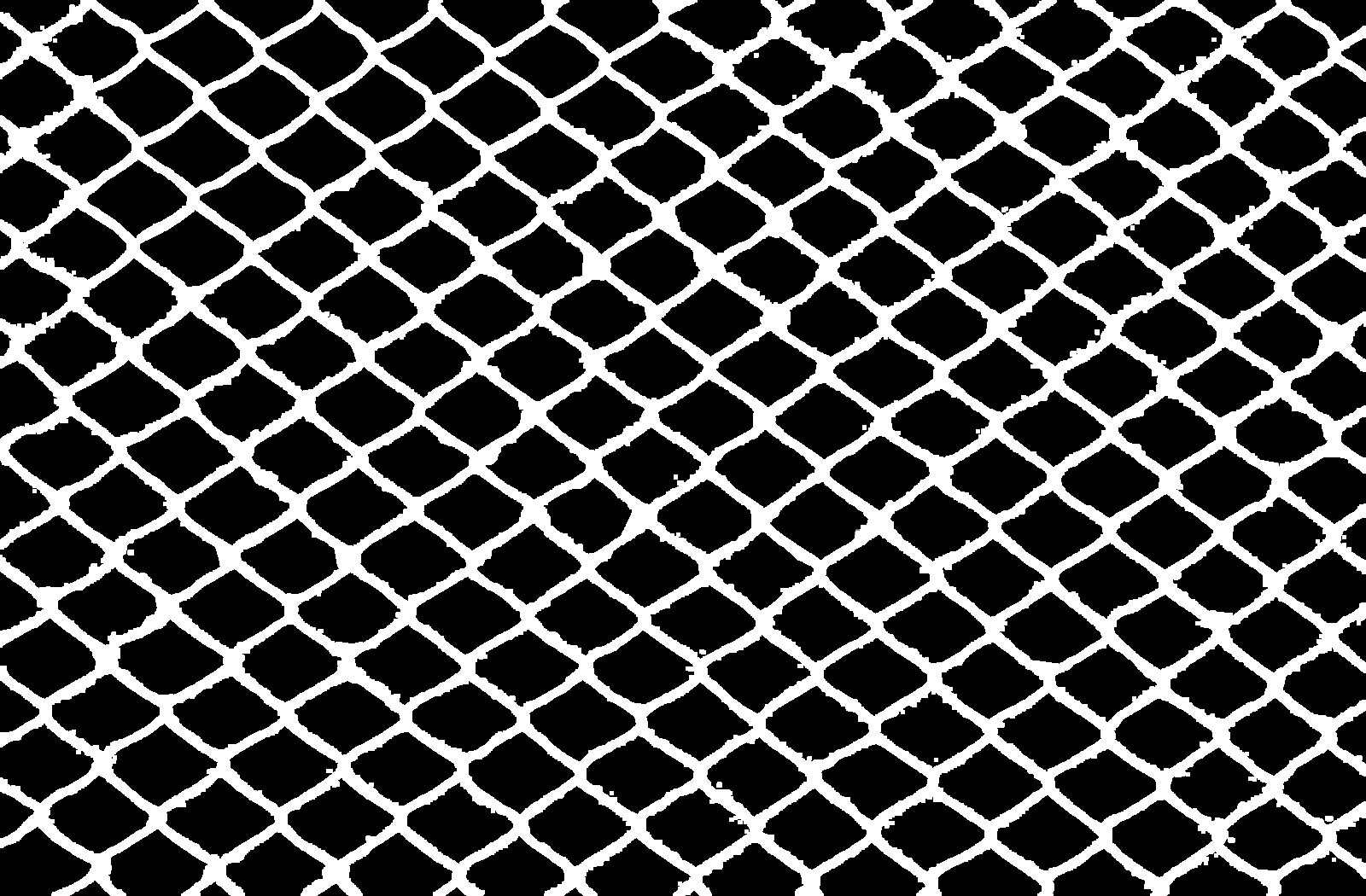}&
		\includegraphics[width=2.9cm]{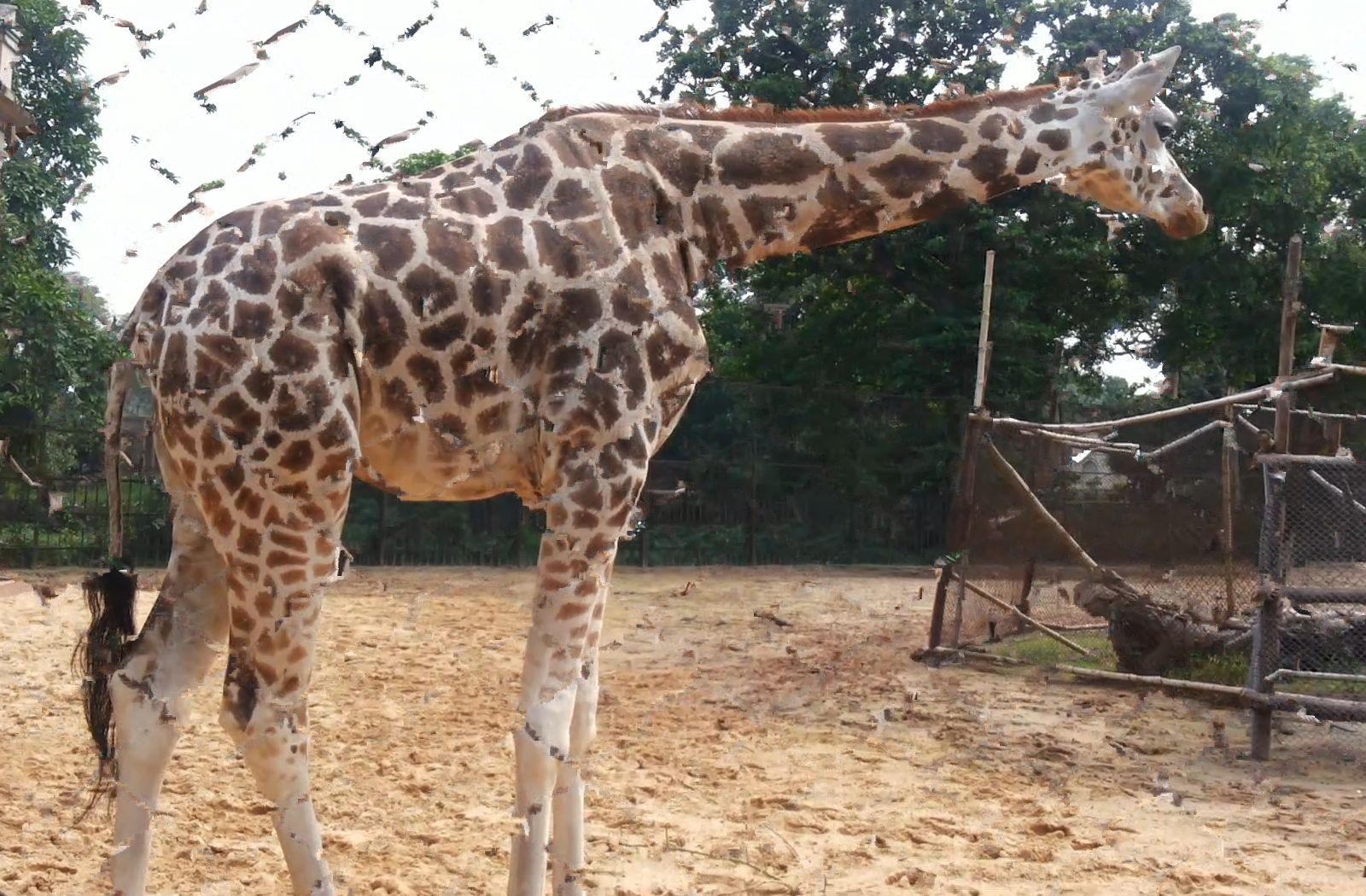}&
		\includegraphics[width=2.9cm]{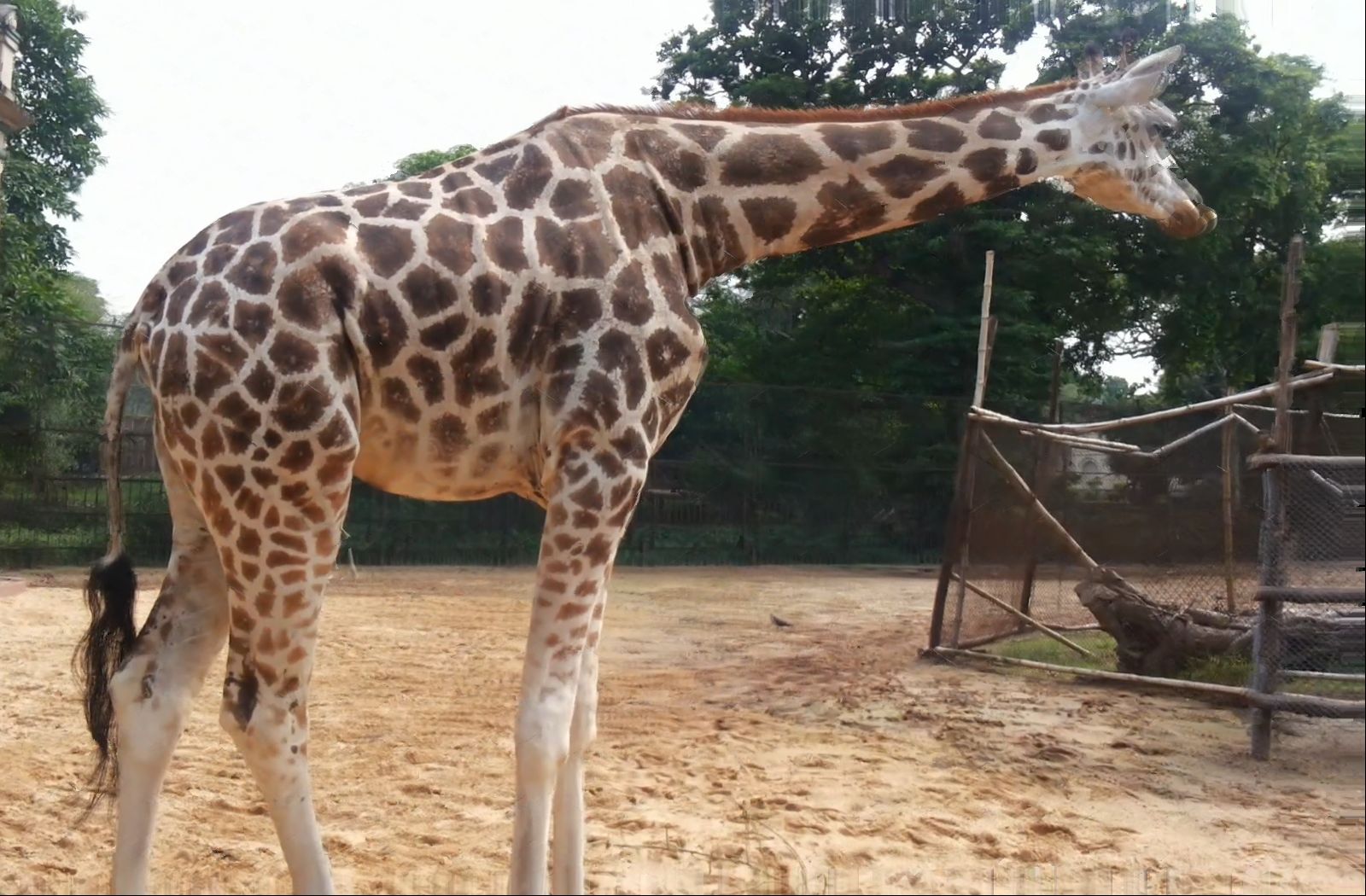}\\
		(a) & (b) & (c) & (d)\\
	\end{tabular}
	\label{fig:fig1}
	\caption{ (a) A frame taken from a video. (b) Segmented binary fence mask obtained using proposed CNN-SVM algorithm. (c) Inpainted image corresponding to (a) using the method of \cite{Criminisi}. (d) De-fenced image corresponding to (a) using the proposed algorithm.}
\end{figure*}

Images containing fences/occlusions occur in several situations such as photographing statues in museums, animals in a zoo etc. Image de-fencing involves the removal of fences or occlusions in images. De-fencing a single photo is strictly an image inpainting problem which uses data in the regions neighbouring fence pixels in the frame for filling-in occlusions. The works of \cite{Bertalmio,Criminisi,James,Konstantinos} addressed the image inpainting problem wherein a portion of the image which is to be inpainted is specified by a mask manually. As shown in Fig. \ref{fig:fig1} (a), in the image de-fencing problem it is difficult to manually mark all fence pixels since they are numerous and spread over the entire image. The segmented binary fence mask obtained using the proposed algorithm is shown in Fig. \ref{fig:fig1} (b). These masks are used in our work to aid in occlusion-aware optical flow computation and background image reconstruction. In Fig. \ref{fig:fig1} (c), we show the inpainted image corresponding to Fig. \ref{fig:fig1} (a) obtained using the method of \cite{Criminisi}. 
The de-fenced image obtained using the proposed algorithm is shown in Fig. \ref{fig:fig1} (d). As can be seen from Fig. \ref{fig:fig1} (c), image inpainting does not yield satisfactory results when the image contains fine textured regions which have to be filled-in. However, using a video panned across a fenced scene can lead to better results due to availability of additional information in the adjacent frames. 

Although, there has been significant progress in the area of lattice detection \cite{Minwoo,CVPR_2016} and restoration of fenced images/videos \cite{CVPR_2016,Yanxi_cvpr,Vrushali,Yadong,acm_sigg2015}, segmentation of fence or occlusion from a single image and de-fencing scenes containing dynamic elements are still challenging problems. Most of the existing works assume global motion between the frames and use images of static scene elements only \cite{Vrushali,Yadong,acm_sigg2015}. Initial work related to image de-fencing has been reported by Liu et al. \cite{Yanxi_cvpr}, wherein fence patterns are segmented via spatial regularity and the fence occlusions are filled-in using an inpainting algorithm \cite{Criminisi}. Recent attempts for image de-fencing \cite{Yadong,acm_sigg2015} use the parallax cue for fence pattern segmentation using multiple frames from a video. However, these works \cite{Yadong,acm_sigg2015} constrain the scene elements to be static. Another drawback of \cite{Yadong} is that if the scene does not produce appreciable depth parallax fence segmentation is inaccurate. A very recent image de-fencing algorithm \cite{CVPR_2016} exploits both color and motion cues for automatic fence segmentation from dynamic videos.

The proposed algorithm for image de-fencing uses a video captured by panning a camera relative to the scene and requires the solution of three sub-problems. The first task is automatic segmentation of fence pixels in the frames of the captured video. Importantly, unlike existing works \cite{CVPR_2016,Yanxi_cvpr,Vrushali,Yadong,acm_sigg2015}, we propose a machine learning algorithm to segment fences in a \textit{single} image. We propose to use a pre-trained convolutional neural network (CNN) for fence texel joint detection to generate automatic scribbles which are fed to an image matting \cite{Yuanjie} technique to obtain the binary fence mask. Note that sample portions of images marked with yellow colored squares shown in Fig. \ref{fig:fig1} (a) are treated as fence texels in this work. To the best of our knowledge, we are the first to detect fence texels using a pre-trained CNN coupled with an SVM classifier. Secondly, we estimate the pixel correspondence between the reference frame and the additional frames using a modified optical flow algorithm which incorporates the knowledge of location of occlusions in the observations. It is to be noted that existing optical flow algorithms find the relative shift only between pixels \textit{visible} in two frames. Accurate registration of the observations is critical in de-fencing the reference image since erroneous pixel matching would lead to incorrect data fusion from additional frames. The basic premise of our work is that image regions occluded by fence pixels in the reference frame are rendered visible in other frames of the captured video. Therefore, we propose an occlusion-aware optical flow method using fence pixels located in the first step of our image de-fencing pipeline to accurately estimate background pixel correspondences even at occluded image regions. Finally, we fuse the information from additional frames in order to uncover the occluded pixels in the reference frame using an optimization framework. Since natural images are sparse, we use the fast iterative shrinkage thresholding algorithm (FISTA) to solve the resulting ill-posed inverse problem assuming $l_{1}$ norm of the de-fenced image as the regularization prior.

\section{Prior Work}

The problem of image de-fencing has been first addressed in \cite{Yanxi_cvpr}
by inpainting fence pixels of the input image. The algorithm proposed in \cite{Minwo} used multiple images for de-fencing, which significantly improves the performance due to availability of occluded image data in additional frames. The work of \cite{Minwo} used a deformable lattice detection method proposed in \cite{Minwoo} for fence detction. Unfortunately, the method of \cite{Minwoo} is not a robust approach and fails for many real-world images. Khasare et al. \cite{Vrushali} proposed an improved multi-frame de-fencing technique by using loopy belief propagation. However, there are two issues with their approach. Firstly, the work in \cite{Vrushali} assumed that motion between the frames is global. This assumption is invalid for more complex dynamic scenes where the motion is non-global. Also, the method of \cite{Vrushali} used an image matting technique proposed by \cite{Yuanjie} for fence segmentation which involves significant user interaction.
A video de-fencing algorithm \cite{Yadong}, proposed a soft fence segmentation method where visual parallax serves as the cue to distinguish fences from the unoccluded pixels.  Recently, Xue et al. \cite{acm_sigg2015} jointly estimated the foreground masks and obstruction-free images using five frames taken from a video. Apart from the image based techniques, Jonna et al. \cite{Jonna} proposed a multimodal approach for image de-fencing wherein they have extracted the fence masks with the aid of depth maps corresponding to the color images obtained using the Kinect sensor. Very recently, our works \cite{Jonna_ACPR,Jonna_JOSA} addresses the image de-fencing problem. However, the drawback of both the methods \cite{Jonna_ACPR,Jonna_JOSA} is that they do not estimate occlusion-aware optical flow for data fusion.

The proposed algorithm for image de-fencing addresses some of the issues with the existing techniques. Firstly, we propose a machine learning algorithm using CNN-SVM for fence segmentation from a \textit{single} image unlike existing works \cite{CVPR_2016,Yadong,acm_sigg2015}, which need a few frames to obtain the fence masks. Importantly, unlike the works of \cite{Yadong,acm_sigg2015}, the proposed algorithm does not assume that the scene is static but we can handle scenes containing dynamic elements. For this purpose, we propose a modified optical flow algorithm for estimation of pixel correpondence between the reference frame and additional frames after segmenting occlusions. 

\section{Methodology}

We relate the occluded image to the original de-fenced image using a degradation model as follows,
\begin{equation}
\mathbf{O}_{m}\textbf{y}_{m} = \mathbf{y}_{m}^{obs} = \textbf{O}_{m}[\textbf{F}_{m}\textbf{x} + \textbf{n}_{m}]
\end{equation}
where $\textbf{y}_{m}$ are observations containing fences obtained from the captured video, $\textbf{O}_{m}$ are the binary fence masks, $\textbf{F}_{m}$ models the relative motion between frames, $\textbf{x}$ is the de-fenced image and $\textbf{n}_{m}$ is Gaussian noise. As described in section 1, the problem of image de-fencing was divided into three sub-problems, which we elaborate upon in the following sub-sections. 

\subsection{Pre-trained CNN-SVM for fence texel joint detection}

The important property of most outdoor fences is their symmetry about the fence texel joints. Referring to Fig. \ref{fig:fig1} (a), we observe that fence texels appear repetitively throughout the entire image. Convolutional neural nets (CNN), originally proposed by \cite{Lecun_1998}, can be effectively trained to recognize objects directly from images with robustness to scale, rotation, translation, noise etc.  Recently, Krizhevsky et al. \cite{ImageNet_NIPS2012} proved the utility of CNNs for object detection and classification in the ILSVRC challenge \cite{ImageNet_cvpr09}. Since real-world fence texels exhibit variations in color, shape, noise, etc., we are motivated to use CNNs for segmenting these patterns robustly.

Convolutional neural networks belong to a class of deep learning techniques which operate directly on an input image extracting features using a cascade of convolutional, activation and pooling layers to finally predict the image category. The key layer in CNN is the convolutional layer whose filter kernels are learnt automatically via backpropagation. The commonly used non-linear activation functions are sigmoid, tanh, rectified linear unit (ReLU) and maxout \cite{maxout} etc. The pooling layers sub-sample the input data. Overfitting occurs in neural networks when the training data is limited. Recently, a technique called Dropout \cite{Dropout} has been proposed which can improve the generalization capability of CNNs by randomly dropping some of the neurons.

\begin{figure}
	\centering
	\begin{tabular}{c}
		\includegraphics[width=9cm]{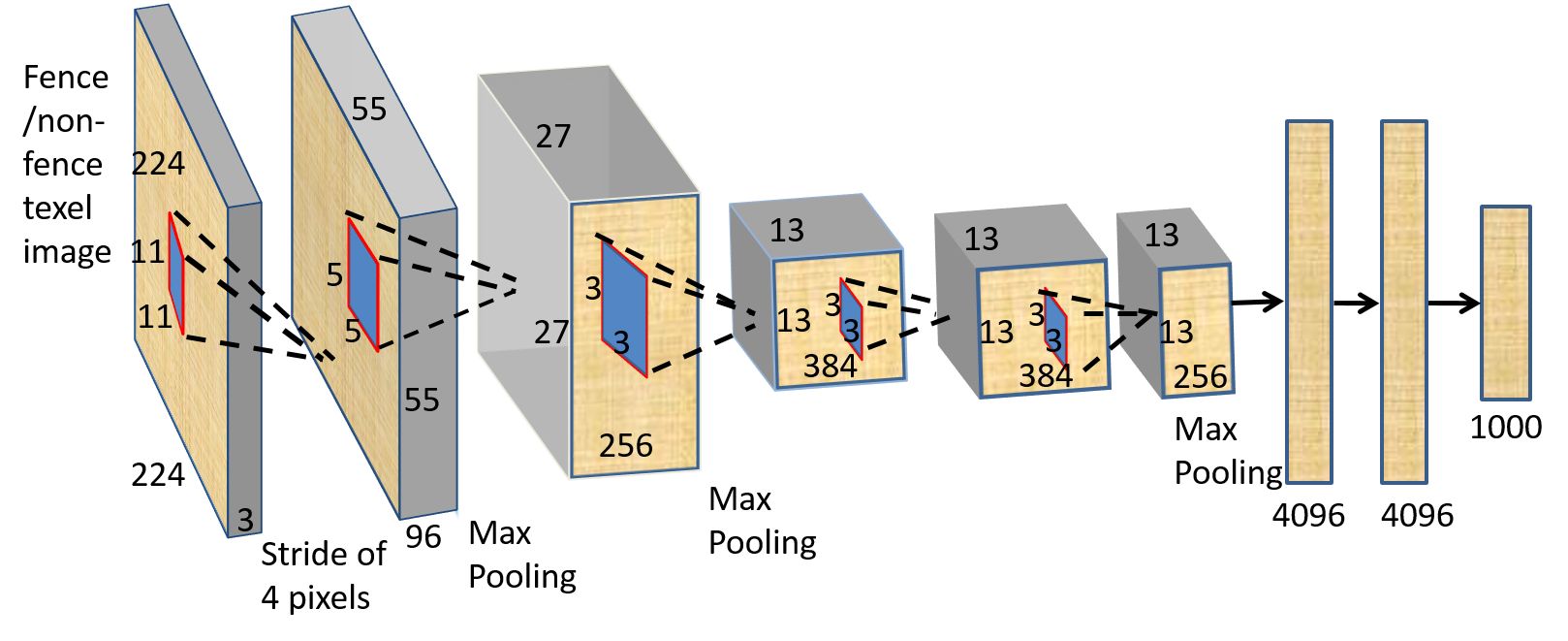}\\
	\end{tabular}
	\caption{The architecture of the pre-trained CNN \cite{ImageNet_NIPS2012}.}
	\label{fig:archi}
\end{figure}

However, since CNNs use supervised learning they need huge labeled datasets and long training time. A possible solution to this problem is to use transfer learning \cite{DeCAF2013,MatConvNet2014}, wherein pre-trained models are used to initialize the weights and fine-tune the network on a different dataset. One can also preserve the pre-trained filter kernels and re-train the classifier part only. In this work, we used a CNN pre-trained on ImageNet \cite{ImageNet_cvpr09} as a feature extractor by excluding the softmax layer. The architecture of the CNN in Fig. \ref{fig:archi} trained on ImageNet contains five convolutinal layers followed by three fully-connected layers and a softmax classifier. Max-pooling layers follow first, second and fifth convolutional layer. 

\begin{figure}[!htb]
	\centering
	\begin{tabular}{c c c}
		\includegraphics[width=4cm]{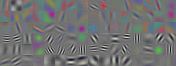} &
		\includegraphics[width=4cm]{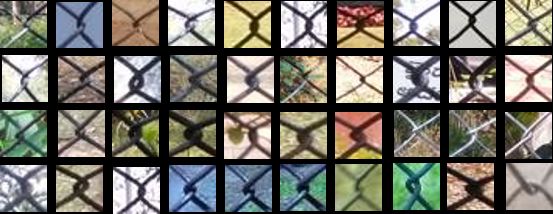} &
		\includegraphics[width=4cm]{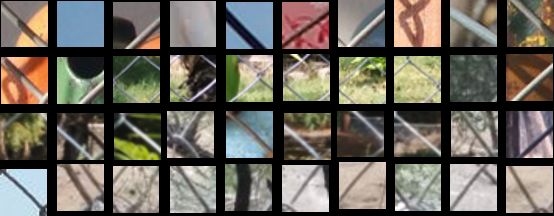}\\
		(a) & (b) & (c)\\
	\end{tabular}
	\caption{(a) $96$ learned filter kernels of size
		$11 \times 11 \times 3$ extracted from the first convolutional layer. (b) Sample fence texel joints. (c) Examples of non-fence texel joints.}
	\label{fig:conv1}
\end{figure}

In Fig. \ref{fig:conv1} (a), we show the $96$ filter kernels of dimensions $11\times 11\times 3$ learned by the first convolutional layer on input images. In this work, we propose to use CNN as a generic feature extractor followed by a support vector machine classifier (CNN-SVM). A given RGB input image is resized to $224\times 224 \times 3$ and fed to the proposed CNN-SVM a feature vector of size $4096$ is extracted from the seventh fully-connected layer.

\begin{figure}
	\centering
	\begin{tabular}{c}
		\includegraphics[width=10cm]{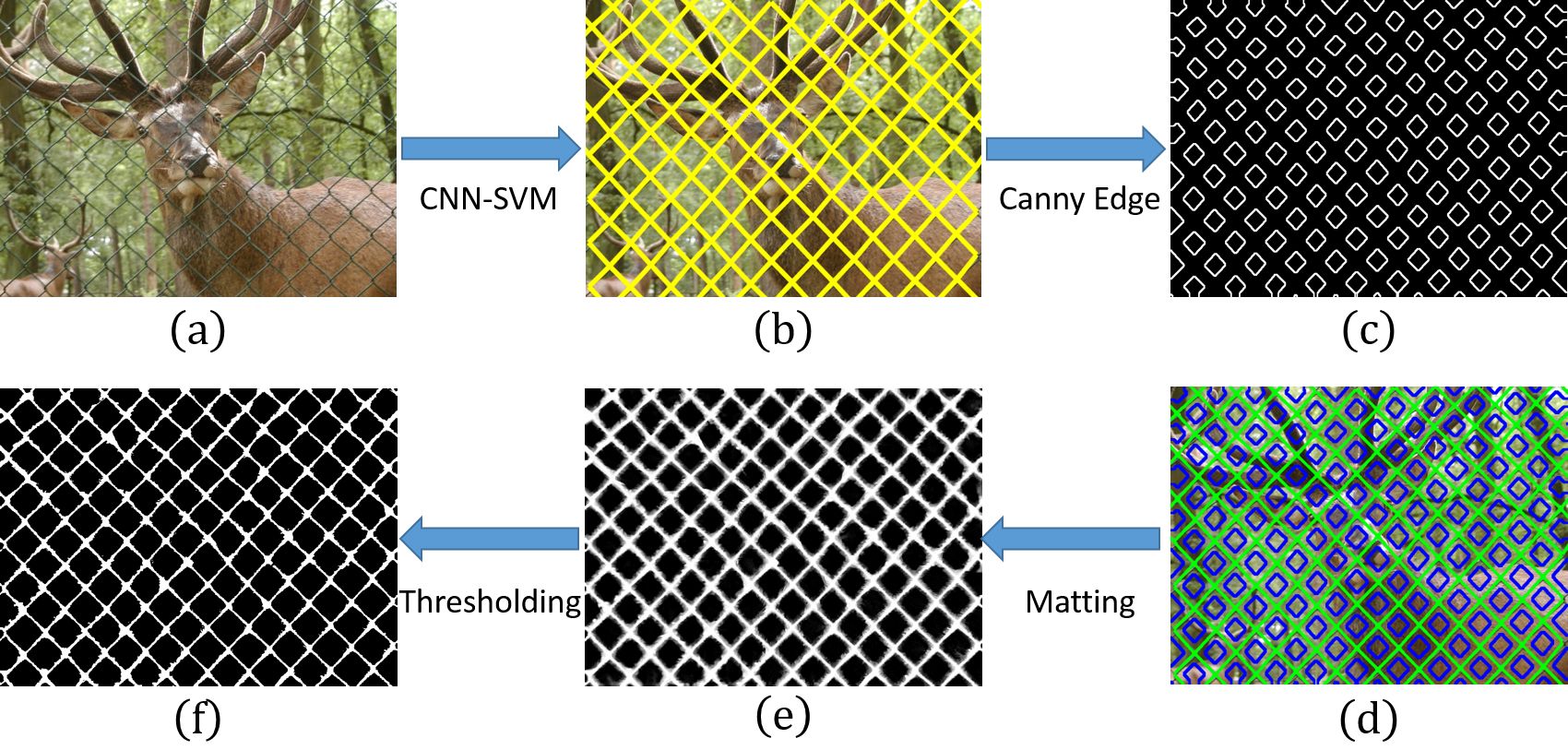}\\
	\end{tabular}
	\caption{Schematic of fence mask segmentation.}
	\label{fig:seg_flow}
\end{figure}

An SVM classifier has been trained to detect fence texels using on these features of dimension $4096$ extracted by the pre-trained CNN from a dataset of $20,000$ fence texel joints and $40,000$ non-fence texel sub-images. In Figs. \ref{fig:conv1} (b) and (c), we show samples of fence texel texels and non-fence texels, respectively. During the testing phase, a sliding window is used to densely scan the test image shown in Fig. \ref{fig:seg_flow} (a) from left to right and top to bottom with a stride of $5$ pixels. The overall workflow of the proposed fence segmentation algorithm is shown in Fig. \ref{fig:seg_flow}. Detected fence texels are joined by straight edges as shown in Fig. \ref{fig:seg_flow} (b). In Fig. \ref{fig:seg_flow} (c) we show the response obtained by Canny edge detection \cite{Canny} algorithm after dilating the preliminary fence mask shown in Fig. \ref{fig:seg_flow} (b) and treated as background scribbles. The combination of both foreground and background scribbles is shown in Fig. \ref{fig:seg_flow} (d), wherein foreground scribbles are obtained by erosion operation on the image in Fig. \ref{fig:seg_flow} (b). We fed these automatically generated scribbles to the method of \cite{Yuanjie} and obtain the alpha map in Fig. \ref{fig:seg_flow} (e). Finally, the binary fence mask shown in Fig. \ref{fig:seg_flow} (f) is generated by thresholding the alpha map obtained from \cite{Yuanjie}. 

\subsection{Occlusion aware optical flow}

The image alignment problem becomes more complex when real-world videos contain dynamic objects. Handling motion boundaries and occlusions in videos for optical flow computation is still challenging. Internal occlusions due to the layered dynamic objects and external occlusions such as fences make the problem tougher. In some practical applications of computer vision  such as view synthesis, image de-fencing, etc we need to compute the correspondence of all pixels between two images despite occlusions.
Many algorithms for estimating optical flow are proposed in the literature \cite{Brox,MDOF,Brox_2004,Liu}, which are based on modifications of the basic variational framework proposed by Horn et al. \cite{HS} addressing its various shortcomings. Recently, significant progress has been made in order to compute dense optical flow in a robust manner \cite{MDOF,Epicflow,deepflow_iccv2013}. The state-of-the-art optical flow algorithms \cite{Brox,MDOF} integrate descriptor matching between two images in a variational framework. It is due to a robust function in the variational framework that the algorithm in \cite{Brox} can handle small internal occlusions. However, it fails to tackle large external occlusions. The algorithm of \cite{Epicflow} computes dense correspondence between images by performing sparse-dense interpolation under contour and motion boundary assumption. An occlusion aware optical flow algorithm is proposed by \cite{Ince}, wherein occlusions in images are handled using a three-step procedure. Initially, the method in \cite{Ince} estimates occlusion-ignorant optical flow. Subsequently,  occlusions are computed using this unreliable optical flow. Finally, the algorithm in \cite{Ince} corrects the optical flow using estimated occlusions.

The basic cue behind the proposed image de-fencing algorithm is that occluded image data in the reference frame is uncovered in additional frames of the captured video. Relative motion among observations needs to be estimated to fuse the information uncovered in the additional images for filling in occlusions in the reference frame. State-of-the-art optical flow algorithms estimate the flow of \textit{visible} areas between two images. However, as described above, there are occlusions in images due to depth changes, dynamic scene elements and external hindrances such as fences/barricades.
If we apply the conventional optical flow algorithms to register two images containing fence occlusions we encounter two difficulties while aligning corresponding fence and background pixels. Firstly, large motion discontinuities exist at the spatial location of fences due to abrupt depth changes which corrupt the estimated optical flow. Secondly, it is to be noted that the background pixels hidden behind the fence assume the flow of fence pixels instead of their own ground truth motion.  Hence, in this work we modify the motion associated with fence pixels to that of surrounding background pixel motion in order to reveal the occluded pixel information in the warped adjacent frame.

In this paper, we re-formulate the optical flow algorithm of \cite{CeLiu} to fit our application of image de-fencing. Akin to \cite{CeLiu}, coarse to fine optical flow is estimated using an incremental framework in Gaussian scale-space. Note that we have already obtained the binary fence mask $\mathbf{O}_{m}$ corresponding to the segmented fence pixels in the observation $\textbf{y}_{m}$. We insert this mask $\mathbf{O}_{m}$ as occlusion operator inside the optical flow framework to deal with the motion inaccuracies at fence locations. At the fence locations data cost is assumed to be zero and only smoothness term in Eq. (3) guides optical flow estimation. We assume sparse gradient prior (modeled using $l_{1}$ norm) for both horizontal and vertical velocities. At every scale, the optimized values are up-scaled and used as initial estimate at the next fine scale. 

Suppose $\textbf{w}=[u,v]$ be the current estimate of horizontal and vertical flow fields and $\tilde{y}_{r}, \tilde{y}_{t}$ be the reference and $t^{th}$ adjacent images, respectively. Under the incremental framework \cite{CeLiu,VidSR_PAMI2014}, one needs to estimate the best increment $d\textbf{w}= (du,dv)$ as follows
\begin{equation}
E(du,dv) =  \arg \min_{d\textbf{w}} \parallel \mathbf{F}_{\textbf{w}+d\textbf{w}}\tilde{y}_{t} - \tilde{y}_{r}\parallel_{1} + \mu \parallel \nabla (u+du)\parallel_{1} + \mu\parallel \nabla (v+dv)\parallel_{1}
\end{equation}
where $\mathbf{F}_{\textbf{w}+d\textbf{w}}$ is the warping matrix corresponding to flow $\textbf{w}+d\textbf{w}$, $\nabla$ is the gradient operator and $\mu$ is the regularization parameter. To use gradient based methods, we replace the $l_{1}$ norm with a differentiable approximation $\phi(x^{2})=\sqrt{x^{2}+\epsilon^{2}}$, $\epsilon=0.001$. To robustly estimate optical flow under the known fence occlusions we compute the combined binary mask $\mathbf{O} = \mathbf{F}_{\textbf{w}+d\textbf{w}}\mathbf{O}_{t}||\mathbf{O}_{r}$ obtained by the logical OR operation between the reference fence mask and backwarped fence from the $t^{th}$ frame using warping matrix $\mathbf{F}_{\textbf{w}+d\textbf{w}}$. To estimate the optical flow increment in the presence of occlusions we disable the data fidelity term by incorporating $\mathbf{O}$ in Eq. (2) as
\begin{equation}
E(du,dv) = \arg \min_{d\textbf{w}} \parallel \mathbf{O}(\mathbf{F}_{\textbf{w}+d\textbf{w}}\tilde{y}_{t} - \tilde{y}_{r})\parallel_{1} + \mu \parallel \nabla (u+du)\parallel_{1} + \mu\parallel \nabla (v+dv)\parallel_{1}
\end{equation}
By first-order Taylor series expansion, 

\begin{equation}
\mathbf{F}_{\textbf{w}+d\textbf{w}} \tilde{y}_{t} \approx \mathbf{F}_{\textbf{w}}\tilde{y}_{t} + \mathbf{Y}_{x}du + \mathbf{Y}_{y}dv
\end{equation}
where $\mathbf{Y}_{x} = diag(\mathbf{F}_{\textbf{w}}\tilde{y}_{t_{x}})$, $\mathbf{Y}_{y} = diag(\mathbf{F}_{\textbf{w}}\tilde{y}_{t_{y}})$, $\tilde{y}_{t_{x}} = \frac{\partial}{\partial x} \tilde{y}_{t}$ and $\tilde{y}_{t_{y}} = \frac{\partial}{\partial y} \tilde{y}_{t}$. We can write Eq. (3) as

\begin{equation}
\begin{split}
\arg\min_{d\textbf{w}} \parallel \mathbf{O} \textbf{F}_{\textbf{w}}\tilde{y}_{t} + \mathbf{O}\mathbf{Y}_{x}du + \mathbf{O}\mathbf{Y}_{y}dv-\mathbf{O}\tilde{y}_{r})\parallel_{1} +  \mu \parallel  \nabla  {(u+du)} \parallel_{1} \\ +   \mu \parallel  \nabla  {(v+dv)} \parallel_{1}
\end{split}
\end{equation}

To estimate the best increments $du$, $dv$ to the current flow $u, v$ we equate the gradients $\left[ \frac{\partial E}{\partial du}; \frac{\partial E}{\partial dv} \right ]$ to zero.

\begin{equation*}
\begin{split}
\begin{bmatrix}
\mathbf{Y}_{x}^{T}\mathbf{O}^{T}\mathbf{W}_{d}\mathbf{O}\mathbf{Y}_{x}+\mu L & \mathbf{Y}_{x}^{T}\mathbf{O}^{T}\mathbf{W}_{d}\mathbf{O}\mathbf{Y}_{y}  \\
\mathbf{Y}_{y}^{T}\mathbf{O}^{T}\mathbf{W}_{d}\mathbf{O}\mathbf{Y}_{x} & \mathbf{Y}_{y}^{T}\mathbf{O}^{T}\mathbf{W}_{d}\mathbf{O}\mathbf{Y}_{y}+\mu L
\end{bmatrix}
\begin{bmatrix}
du \\
dv 
\end{bmatrix}\\	
=
\begin{bmatrix}
-Lu-\mathbf{Y}_{x}^{T}\mathbf{O}^{T}\mathbf{W}_{d}\mathbf{O}\mathbf{F}_{\mathbf{w}}\tilde{y}_{t}+\mathbf{Y}_{x}^{T}\mathbf{O}^{T}\mathbf{W}_{d}\mathbf{O}\tilde{y}_{r} \\
-Lv-\mathbf{Y}_{y}^{T}\mathbf{O}^{T}\mathbf{W}_{d}\mathbf{O}\mathbf{F}_{\mathbf{w}}\tilde{y}_{t}+\mathbf{Y}_{y}^{T}\mathbf{O}^{T}\mathbf{W}_{d}\mathbf{O}\tilde{y}_{r}
\end{bmatrix}
\end{split}	
\end{equation*}
where $L=\mathbf{D}_{x}^T\mathbf{W}_{s}\mathbf{D}_{x} + \mathbf{D}_{y}^T\mathbf{W}_{s}\mathbf{D}_{y}$, $\mathbf{W}_{s} = diag(\phi'(|\nabla u|^2))$ and $\mathbf{W}_{d} = diag(\phi'(|\mathbf{O}\\ \mathbf{F}_{\textbf{w}}\tilde{y}_{t}-\mathbf{O}\tilde{y}_{r}|^2))$. We define $\mathbf{D}_{x}$ and $\mathbf{D}_{y}$ are discrete differentiable operators along horizontal and vertical directions, respectively. We used conjugate gradient (CG) algorithm to solve for 		
$d\mathbf{w}$ using iterative re-weighted least squares (IRLS) framework. 

\subsection{FISTA Optimization Framework}
Once the relative motion between the frames has been estimated we need to fill-in the occluded pixels in the reference image using the corresponding uncovered pixels from the additional frames. Reconstructing de-fenced image $\mathbf{x}$ from the occluded observations is an ill-posed inverse problem and therefore prior information for $\mathbf{x}$ has to be used to regularize the solution. Since natural images are sparse, we employed $l_{1}$ norm of the de-fenced image as regularization constraint in the optimization framework as follows,
\begin{equation}
\hat{\textbf{x}} = \arg\min_{\textbf{x}} \left[ \sum_{m}\parallel  \mathbf{y}_{m}^{obs}-\mathbf{O}_{m}\mathbf{F}_{m}\mathbf{x}\parallel^{2} +  \lambda \parallel \mathbf{x} \parallel_{1}  \right]
\end{equation}
where $\lambda$ is the regularization parameter.  

Since the objective function contains $l_{1}$ norm as a regularization function, it is difficult to solve Eq. 6 with the conventional gradient-based algorithms. Here, we employed one of the proximal algorithms such as FISTA \cite{FISTA_SIAM2009} iterative framework to handle non-smooth functions for image de-fencing. The key step in FISTA iterative framework is the proximal operator \cite{Prox_TV1} which operates on the combination of two previous iterates.

\begin{algorithm}[!htb]
	\caption{FISTA image de-fencing}\label{alg:srd}
	\begin{algorithmic}[1]
		\State $\textbf{Input:} \lambda, \alpha, \mathbf{z}_1 = \mathbf{x}_0 \in \mathbb{R}^{M\times N}, t_1=1 $
		\Repeat
		\State $ \mathbf{x}_{k} =  prox_{\alpha}(g)(\mathbf{z}_{k} - \alpha \nabla f(\mathbf{z}_{k}))$
		\State $t_{k+1} = \frac{1+\sqrt{1+4t_{k}^2}}{2}$
		\State $\mathbf{z}_{k+1} = \mathbf{x}_{k} + \left(\frac{t_{k}-1}{t_{k+1}}\right)(\mathbf{x}_{k}-\mathbf{x}_{k-1})$
		\State $ k\gets k+1$
		\Until $(\parallel\mathbf{x}_{k}-\mathbf{x}_{k-1}\parallel_{2}\leq\epsilon)$
	\end{algorithmic}
\end{algorithm}\

The proximal operator is defined as the solution of the following convex optimization \cite{prox}
\begin{equation}
prox_{\alpha}(g)(x) = \arg\min_{y}\{g(y)+\frac{1}{2\alpha}\parallel y-x \parallel^2\}
\end{equation}
If  $g(y)$ is $l_{1}$ norm, then $prox_{\alpha}(g)(x) = max(\left|x\right|-\lambda\alpha,0)sign(x)$.
The gradient for data matching cost $f$ is given as follows 
\begin{equation}
\nabla f(\mathbf{z}) = \sum_{m}  \mathbf{F}^{T}_{m}\mathbf{O}^{T}_{m}(\mathbf{O}_{m}\mathbf{F}_{m}\mathbf{z} - \mathbf{y}_{m}^{obs}) 
\end{equation}

\section{Experimental Results}

Initially, we report both qualitative and quantitative results obtained using the proposed fence segmentation algorithm on various datasets. Subsequently, we show the impact of accounting for occlusions in the incremental flow framework. Finally, we report image de-fencing results obtained with the FISTA optimization framework. To demonstrate the efficacy of the proposed de-fencing system, we show comparison results with state-of-the-art fence segmentation, and de-fencing methods in the literature. We used only three frames from each captured video for all the image de-fencing results reported here using the proposed algorithm. For all our experiments, we fixed $\lambda$ = $0.0005$ in Eq. 6. We ran all our experiments on a $3.4$ GHz Intel Core i$7$ processor with $16$ GB of RAM.

\subsection{Fence Segmentation}

For validating the proposed algorithm for fence segmentation, we have evaluated our algorithm on state-of-the-art datasets \cite{Yadong,acm_sigg2015,NRT}. We also show segmentation results on a proposed fenced image dataset consisting of $200$ real-world images captured under diverse scenarios and complex backgrounds. We  report quantitative results on PSU NRT \cite{NRT} dataset and qualitative results on \cite{Yadong,acm_sigg2015,NRT} datasets. As discussed in section 3.1, we have extracted features from $20,000$ fence, $40,000$ non-fence texel images using a pre-trained CNN to train an SVM classifier. The trained classifier is used to detect joint locations in images via a sliding window protocol. We compare the results obtained using a state-of-the-art lattice detection algorithm \cite{Minwoo} and the proposed algorithm on all the datasets.

\begin{figure}[!htb]
	\begin{tabular}{c c c c}
		\includegraphics[width=3cm]{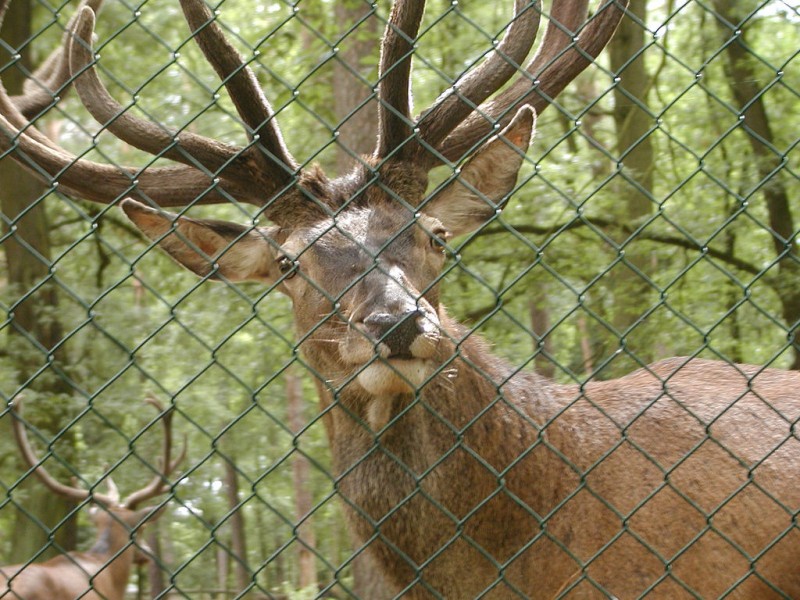}&
		\includegraphics[width=3cm]{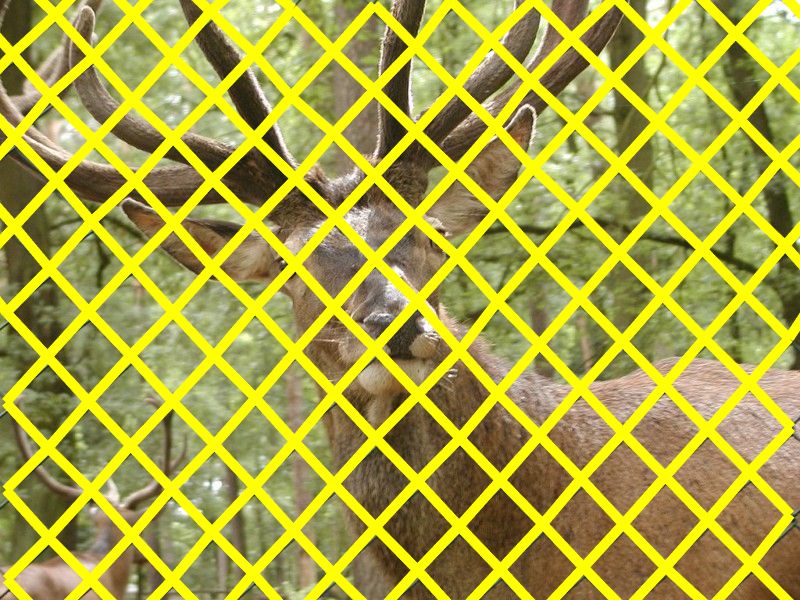}&
		\includegraphics[width=3cm]{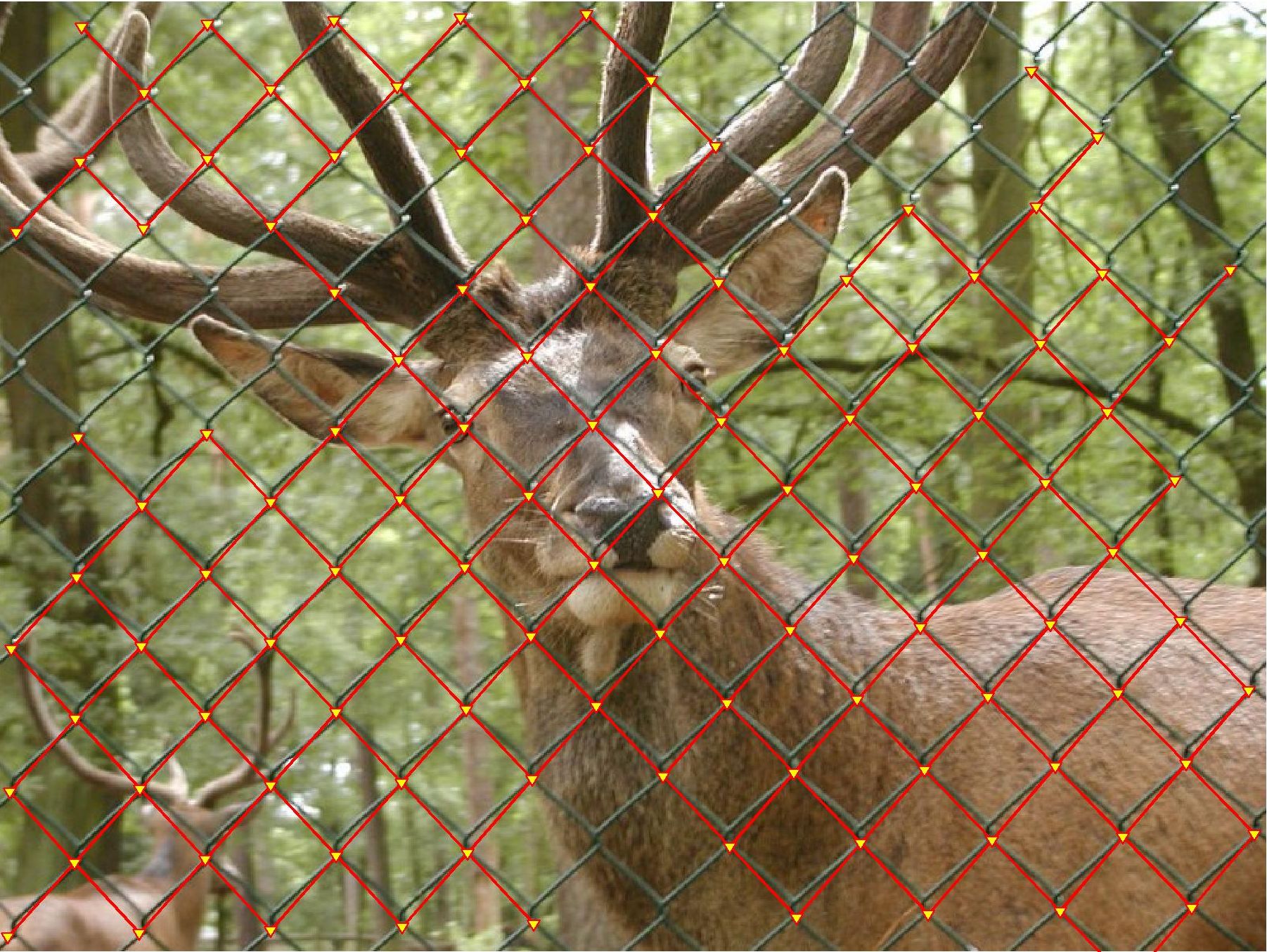}&
		\includegraphics[width=3cm]{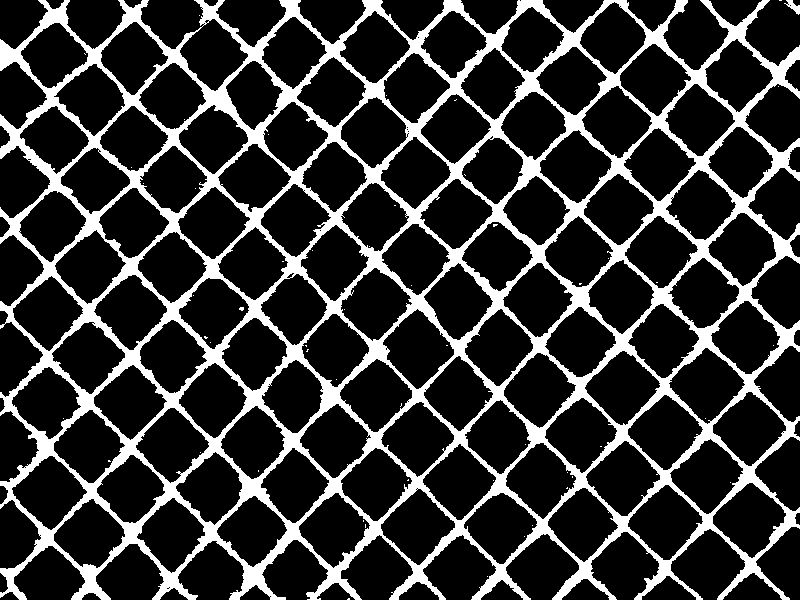}\\
		(a) & (f) & (k) & (p) \\
		\includegraphics[width=3cm]{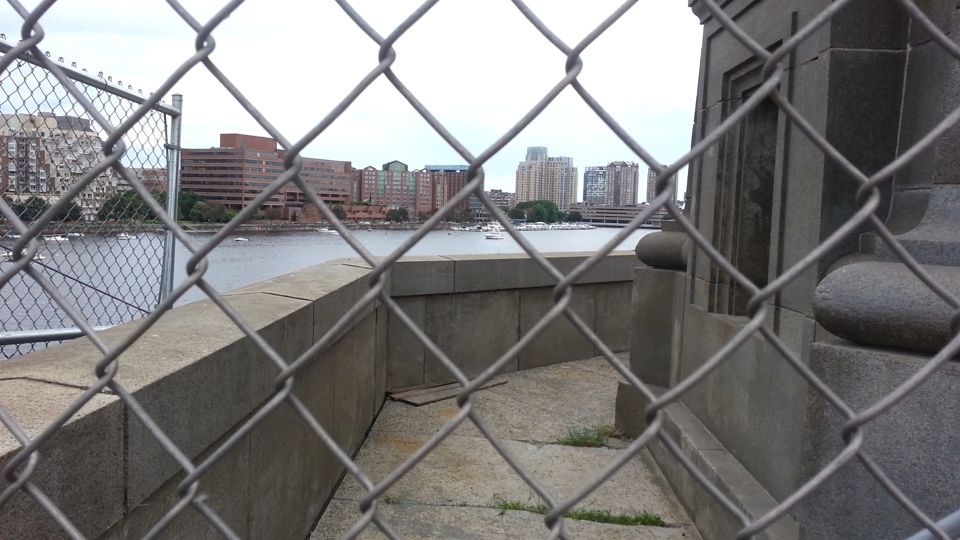}&
		\includegraphics[width=3cm]{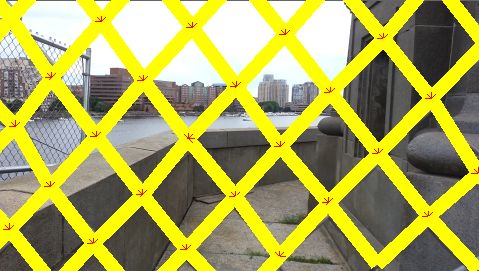}&
		\includegraphics[width=3cm]{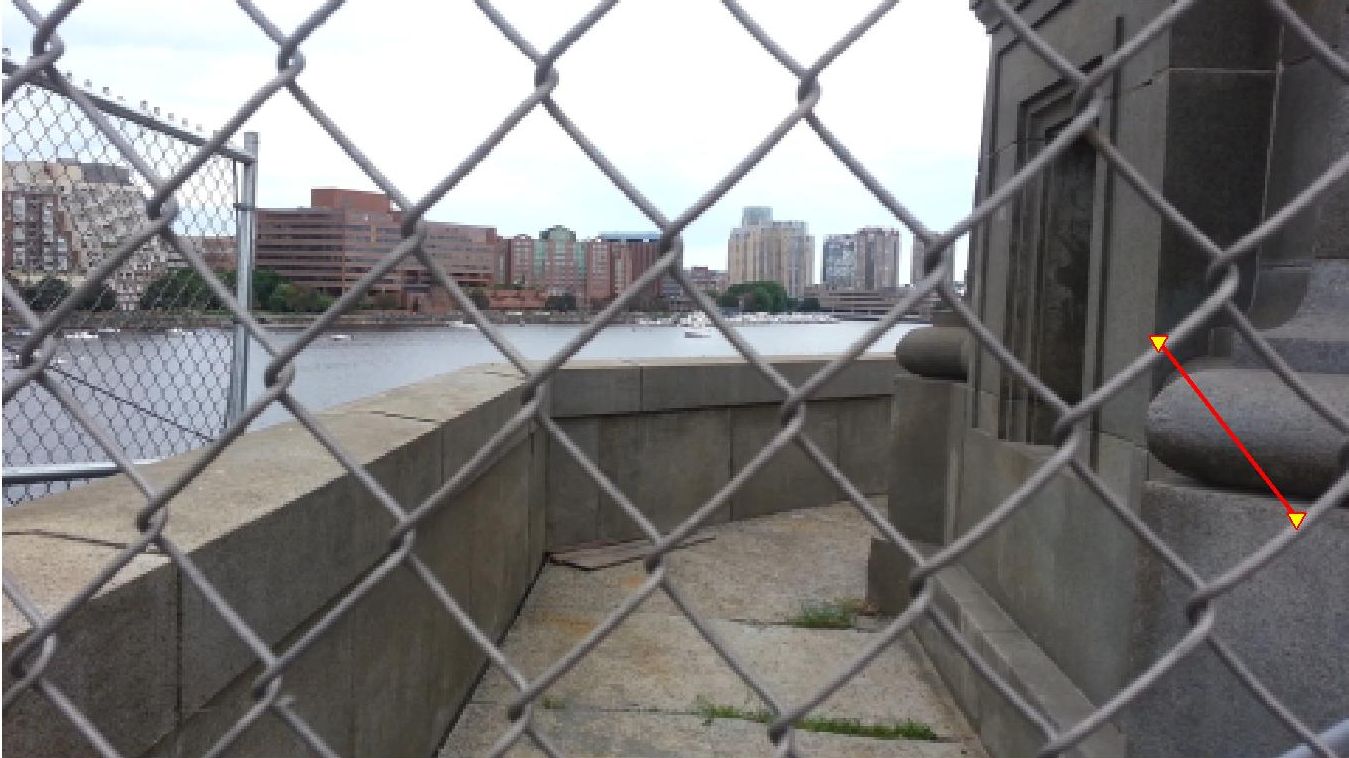}&
		\includegraphics[width=3cm]{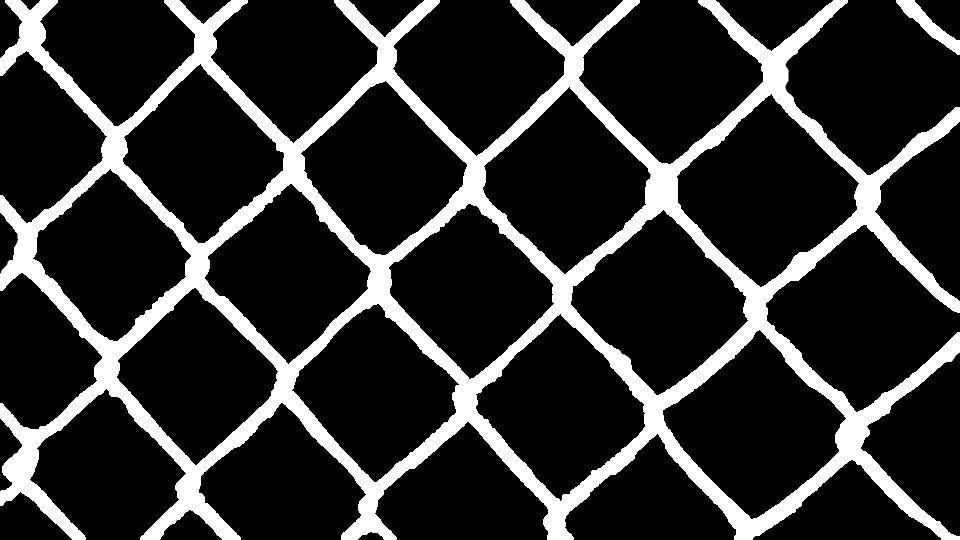}\\
		(b) & (g) & (l) & (q) \\		
		\includegraphics[width=3cm]{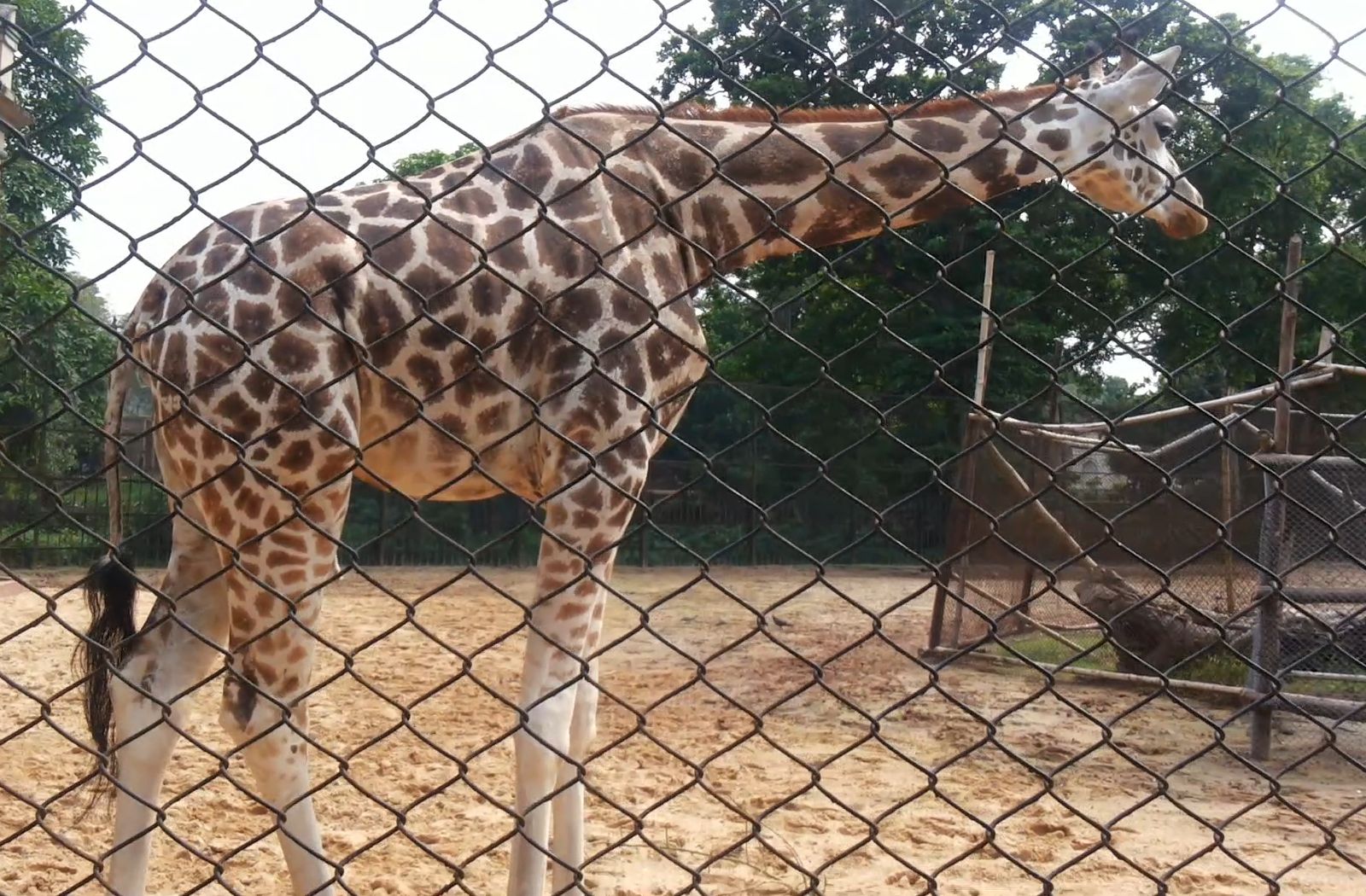}&
		\includegraphics[width=3cm]{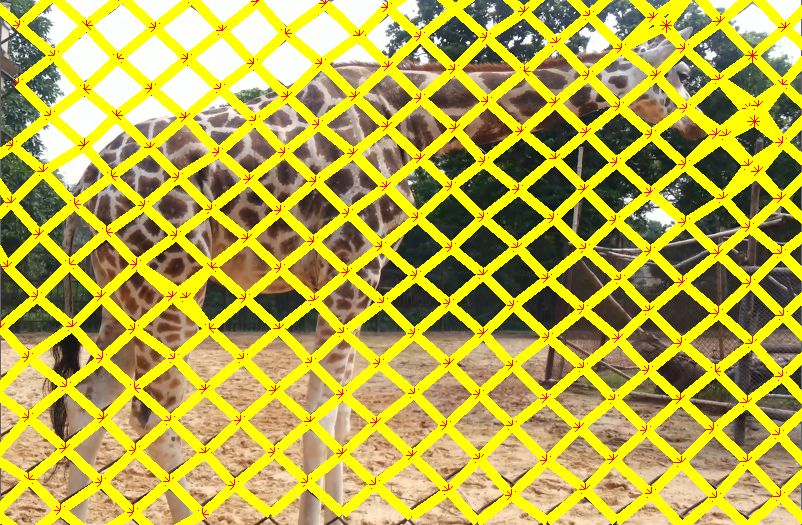}&
		\includegraphics[width=3cm]{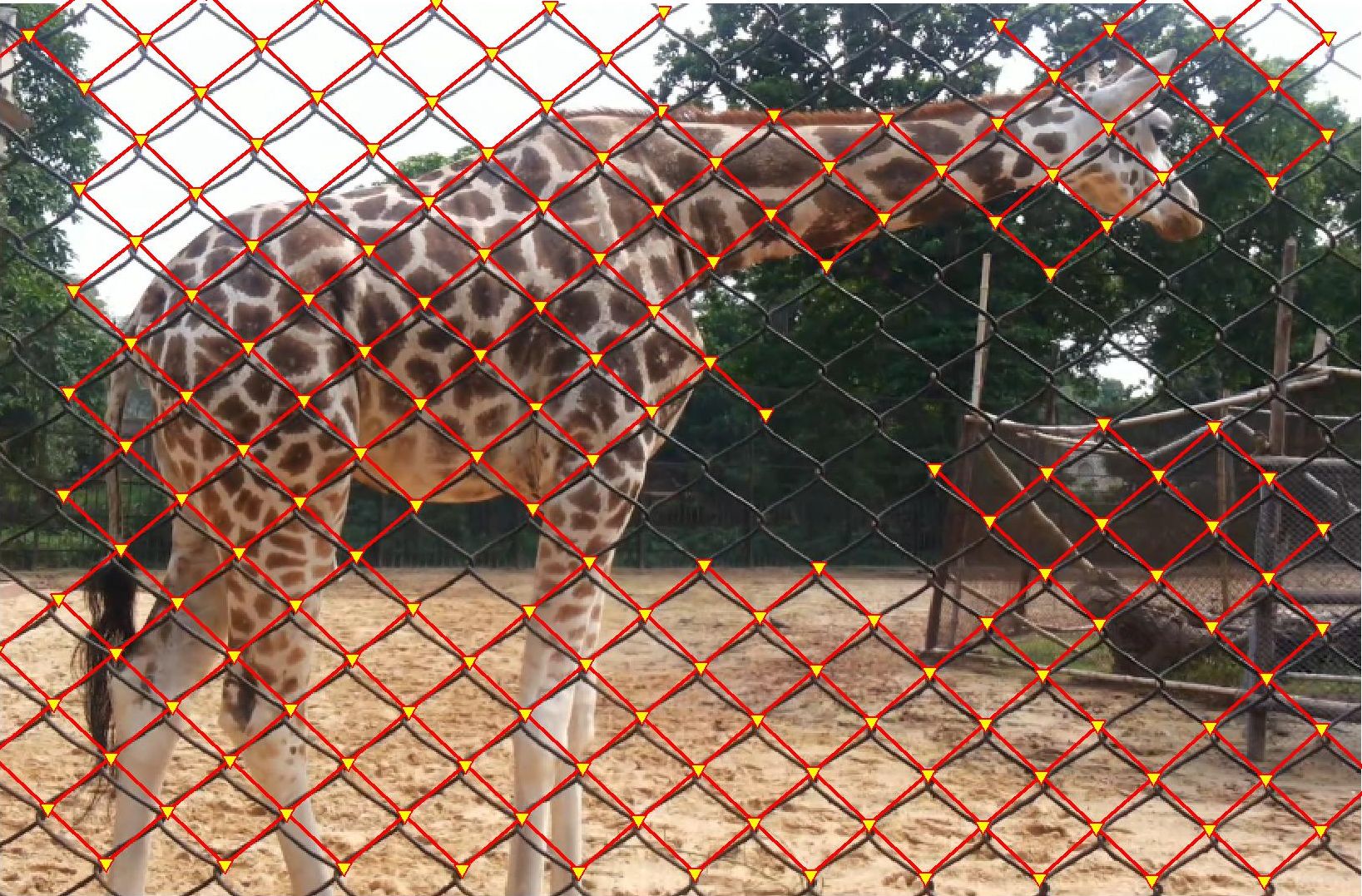}&
		\includegraphics[width=3cm]{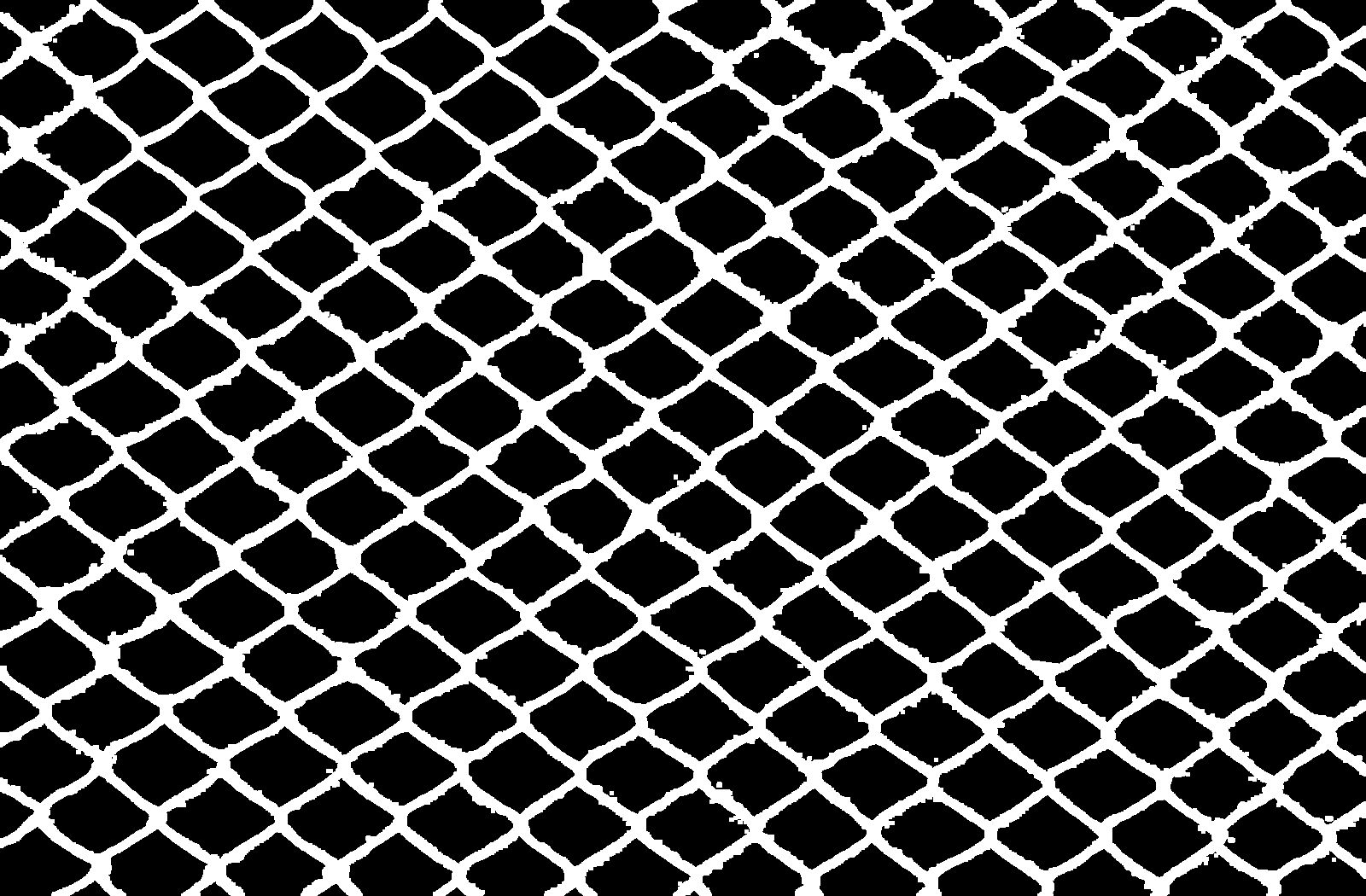}\\
		(c) & (h) & (m) & (r) \\
		\includegraphics[width=3cm]{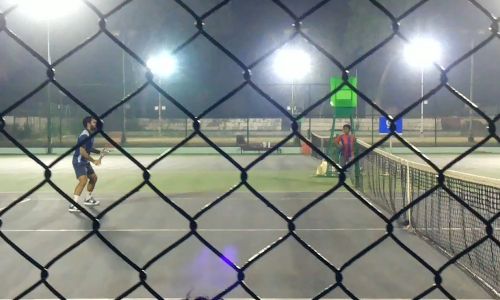}&
		\includegraphics[width=3cm]{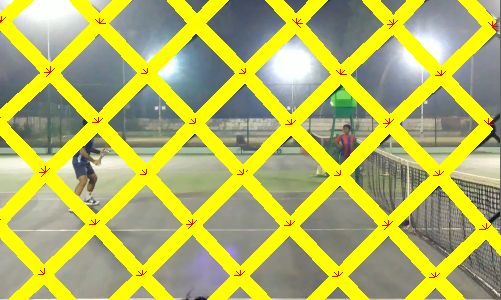}&
		\includegraphics[width=3cm]{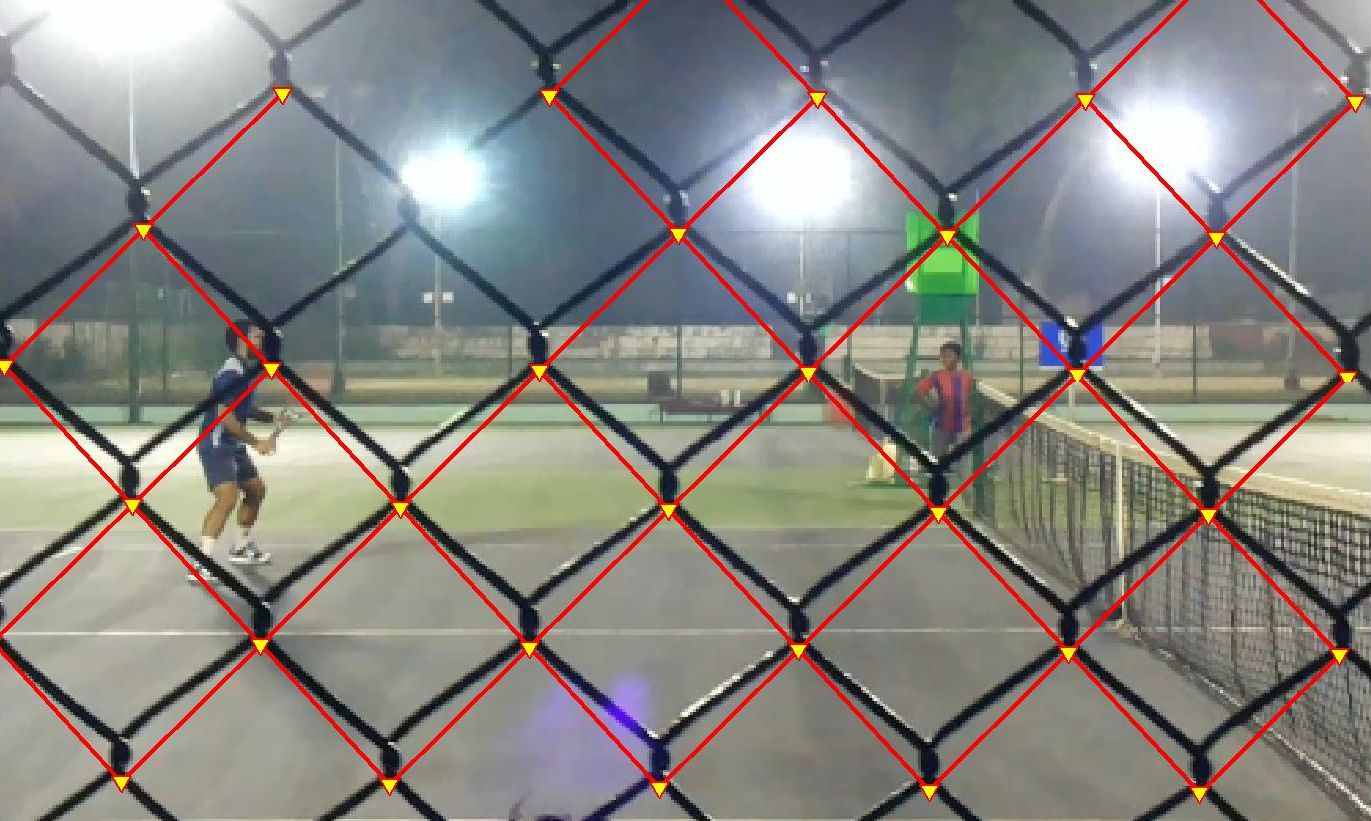}&
		\includegraphics[width=3cm]{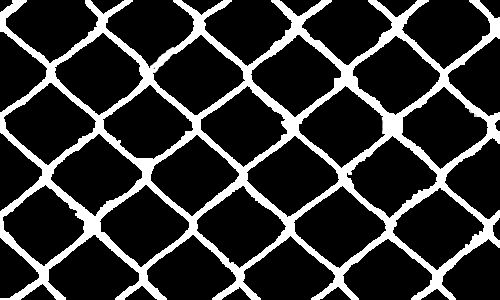}\\
		(d) & (i) & (n) & (s) \\
		\includegraphics[width=3cm]{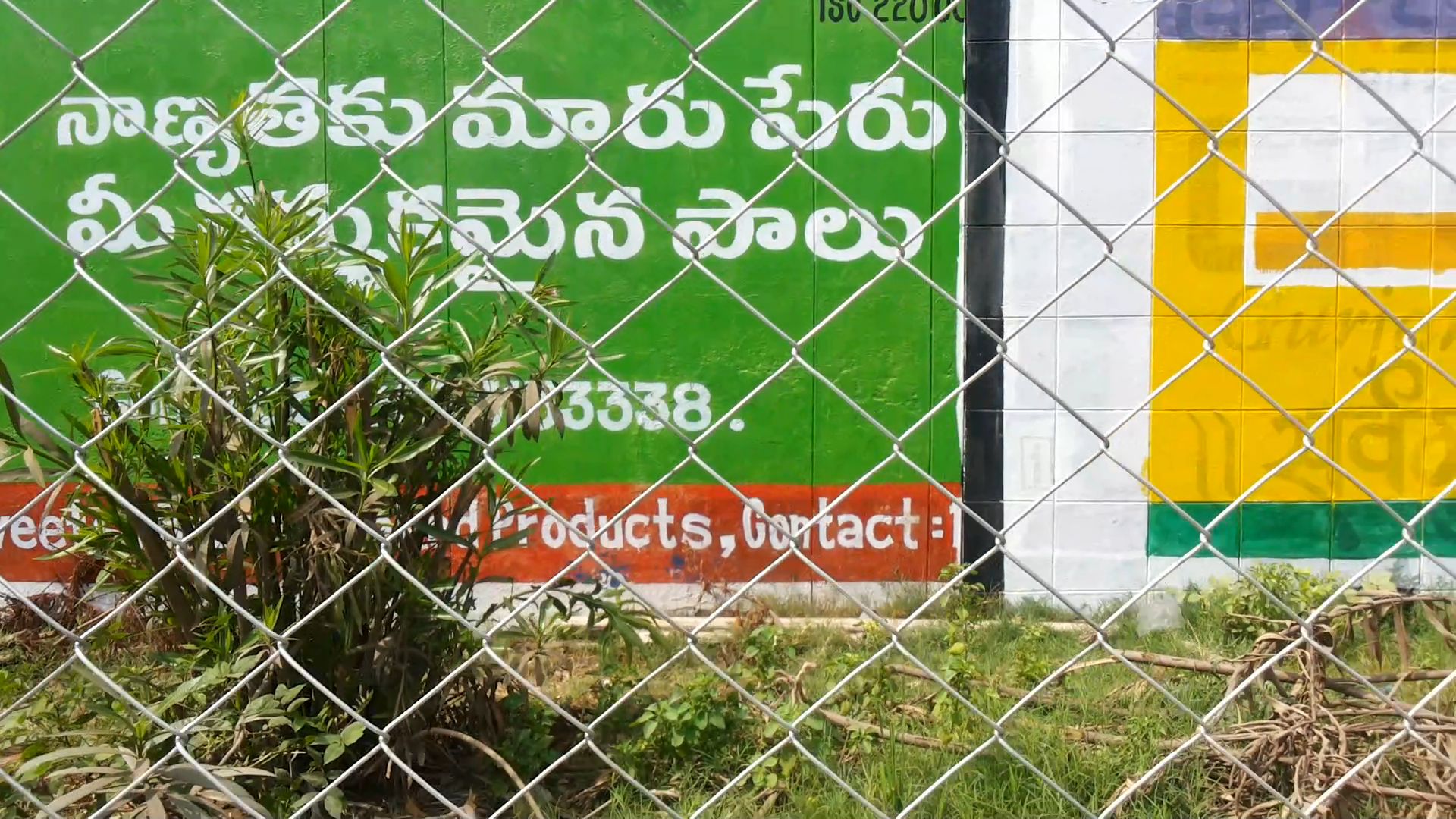}&
		\includegraphics[width=3cm]{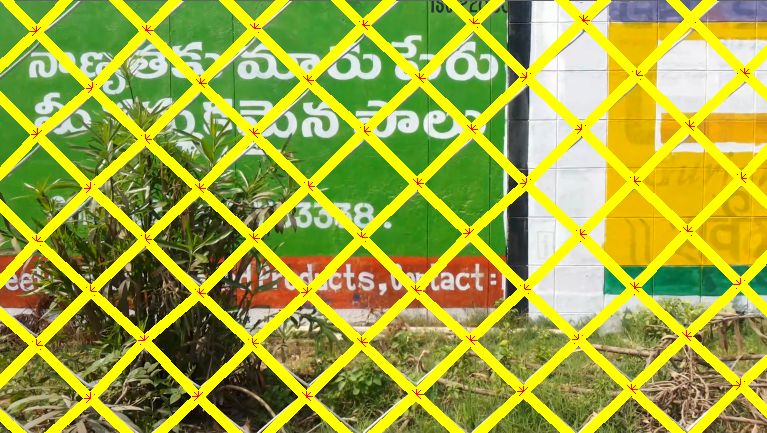}&
		\includegraphics[width=3cm]{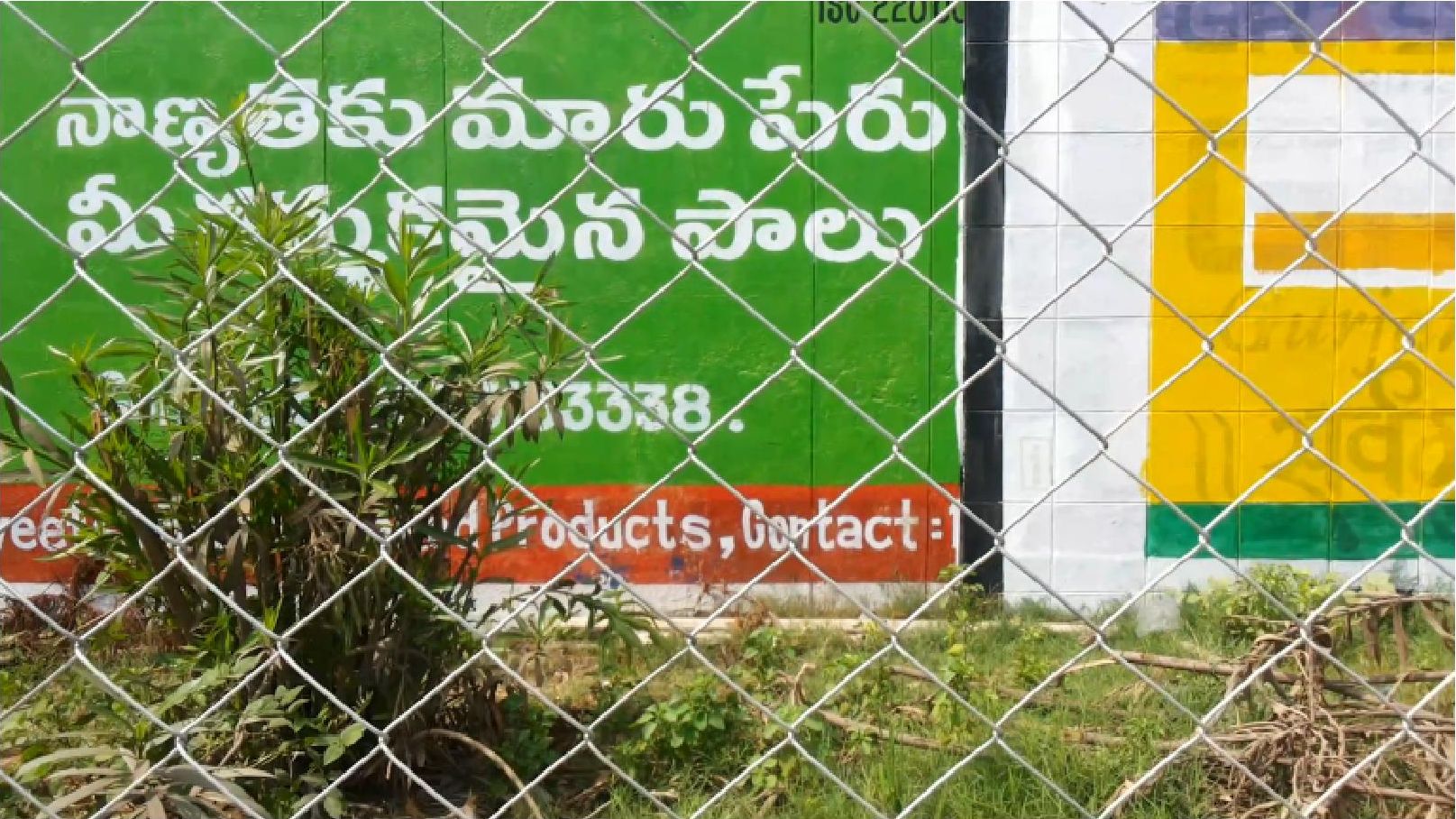}&
		\includegraphics[width=3cm]{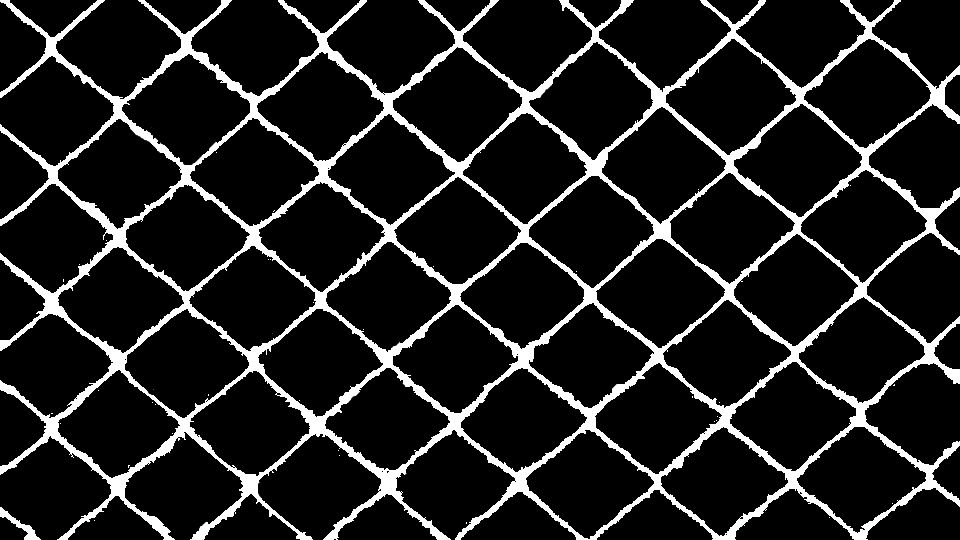}\\
		(e) & (j) & (o) & (t) \\
	\end{tabular}
	\caption{ First column: sample images from NRT \cite{NRT}, \cite{acm_sigg2015} and proposed fenced image datasets, respectively. Second column: fence masks generated using the proposed pre-trained CNN-SVM algorithm. Third column: fence detection using \cite{Minwoo}. Fourth column: final binary fence masks corresponding to images in the first column obtained by generating scribbles using fence detections in images of the  second column which are fed to the method of \cite{Yuanjie}.}
	\label{fig:seg}
\end{figure}

Initially, in Fig. \ref{fig:seg} (a) we show a fenced image from the PSU NRT dataset \cite{NRT}. Fence texels are detected using our pre-trained CNN-SVM approach and are jointed by straight edges, as shown in Fig. \ref{fig:seg} (f). Note that all fence texels are detected accurately in Fig. \ref{fig:seg} (f). In contrast, the method of \cite{Minwoo} failed completely to extract the fence pixels as seen in Fig. \ref{fig:seg} (k). The output of Fig. \ref{fig:seg} (f) is used to generate foreground and background scribbles which are fed to the image matting technique of \cite{Yuanjie}. The final binary fence mask obtained by thresholding the output of \cite{Yuanjie} is shown in Fig. \ref{fig:seg} (p). Next, we have validated both the algorithms on image taken from a recent dataset \cite{acm_sigg2015} shown in Fig. \ref{fig:seg} (b). In Fig. \ref{fig:seg} (g), we show the fence texels detected using our pre-trained CNN-SVM approach and joined by straight edges. In contrast, the method of \cite{Minwoo} failed completely to extract the fence pixels as seen in Fig. \ref{fig:seg} (l). The output of Fig. \ref{fig:seg} (g) is used to generate scribbles as outlined in section 3.1. These foreground and background scribbles are fed to the image matting technique of \cite{Yuanjie}. The final binary fence mask obtained by thresholding the output of \cite{Yuanjie} is shown in Fig. \ref{fig:seg} (q).
Finally, we perform experiments on images from the proposed fenced image dataset. Sample images taken from the dataset are shown in Figs. \ref{fig:seg} (c)-(e). In Figs. \ref{fig:seg} (h)-(j), we show the fence segmentations obtained using the proposed pre-trained CNN-SVM algorithm. We observe that the proposed algorithm detected all the fence texel joints accurately. The lattice detected using \cite{Minwoo} are shown in Figs. \ref{fig:seg} (m)-(o). We can observe that the approach of \cite{Minwoo} partially segments the fence pixels in Fig. \ref{fig:seg} (m). Note that in Fig. \ref{fig:seg} (o) the algorithm of \cite{Minwoo} completely failed to segment fence pixels. The final binary fence masks obtained by thresholding the output of \cite{Yuanjie} are shown in Figs. \ref{fig:seg} (r)- (t). 

A summary of the quantitative evaluation of the fence texel detection method of \cite{Minwoo} and the pre-trained CNN-SVM based proposed algorithm is given in Table 1. The F-measure obtained for \cite{Minwoo} on PSU NRT \cite{NRT} dataset and proposed fenced image datasets are $0.62$ and $0.41$, respectively. In contrast, F-measure for the proposed method on PSU NRT dataset \cite{NRT} and our fenced image datasets are $0.97$ and $0.94$, respectively.

\begin{table}[!htb]
	\centering
	\begin{center}
		\caption{Quantitative evaluation of fence segmentation}
	\end{center}
	\scalebox{0.9}
	{
		\begin{tabular}{|c|c|c|c|c|c|c|}
			\hline
			& \multicolumn{3}{|c|}{NRT Database \cite{NRT}} & \multicolumn{3}{|c|}{Our Database }\\\cline{2-7}
			Method & Precision & Recall & F-measure & Precision & Recall & F-measure\\
			\hline
			Park et al. \cite{Minwoo} & 0.95 & 0.46 & 0.62 & 0.94 & 0.26 & 0.41\\
			\hline
			\textbf{pre-trained CNN-SVM} & 0.96 & 0.98 & \textbf{0.97} & 0.90 & 0.98 & \textbf{0.94}\\
			\hline
		\end{tabular}}
		\label{tab:PerformanceEval1}
	\end{table}
	
	\subsection{Optical Flow under Known Occlusions}
	
\begin{figure}[!htb]
		\centering
		\begin{tabular}{c c c c c}
			\includegraphics[height=1.6cm]{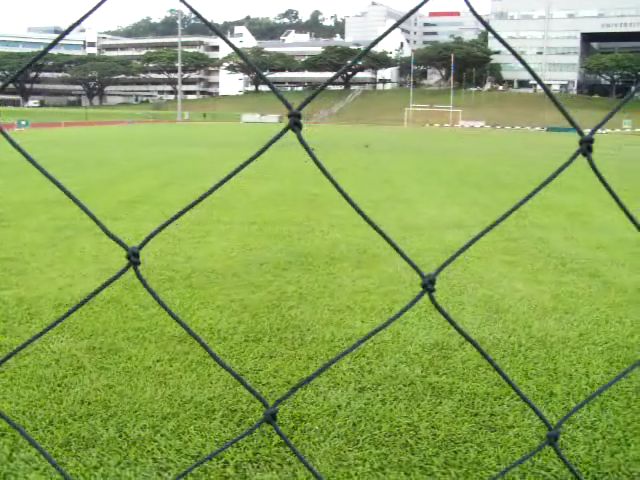}&
			\includegraphics[height=1.6cm]{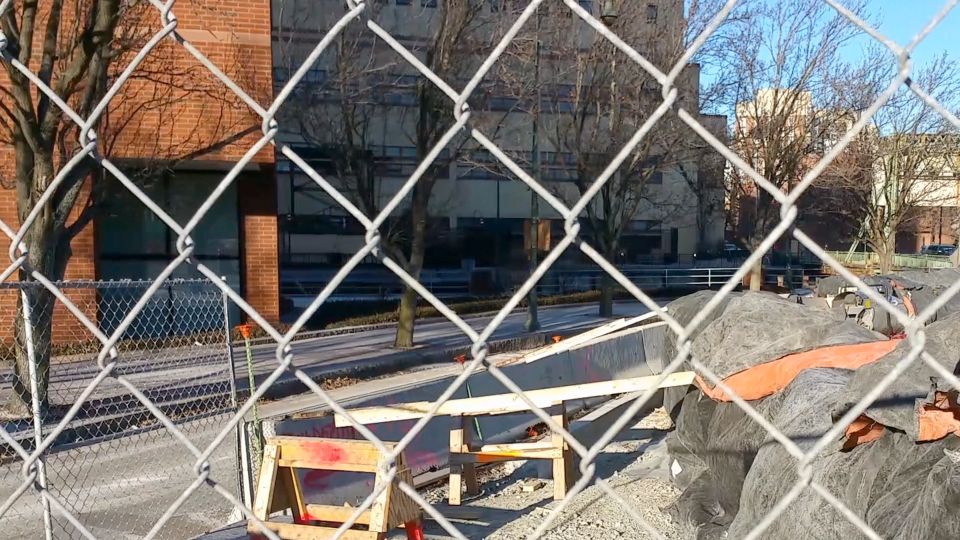}&
			\includegraphics[height=1.6cm]{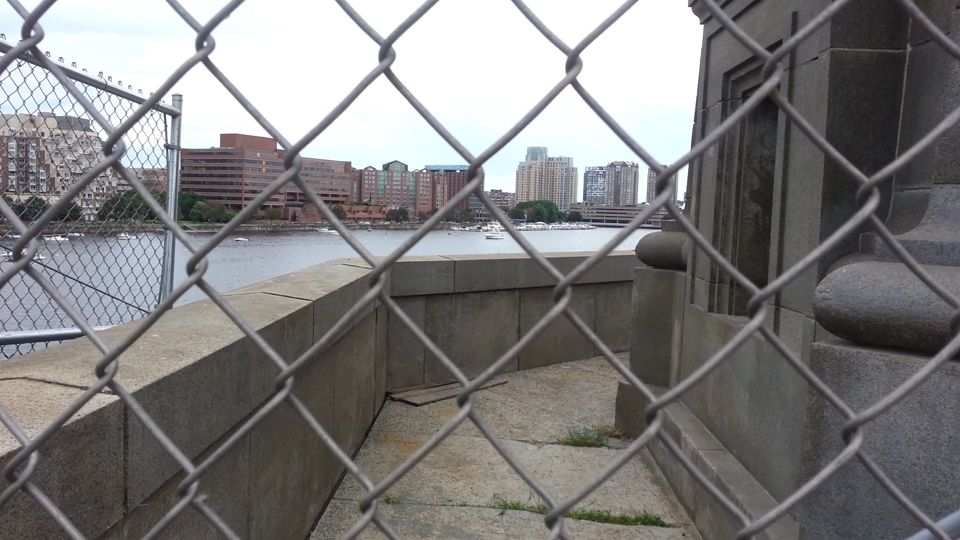}&
			\includegraphics[height=1.6cm]{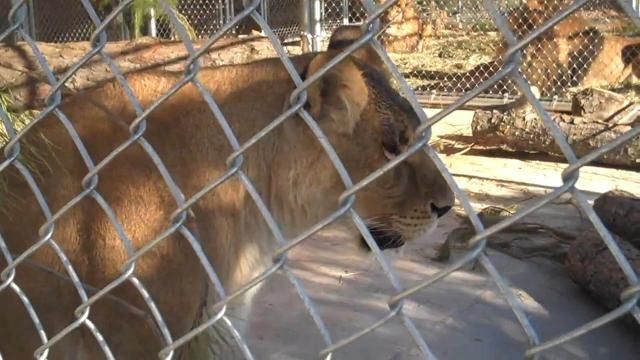}&
			\includegraphics[height=1.6cm]{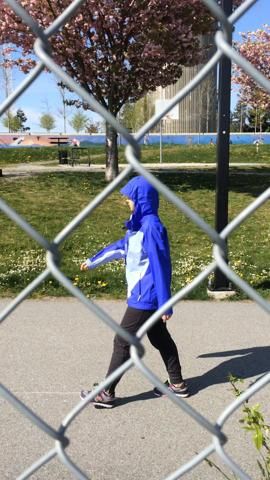}\\
			\includegraphics[height=1.6cm]{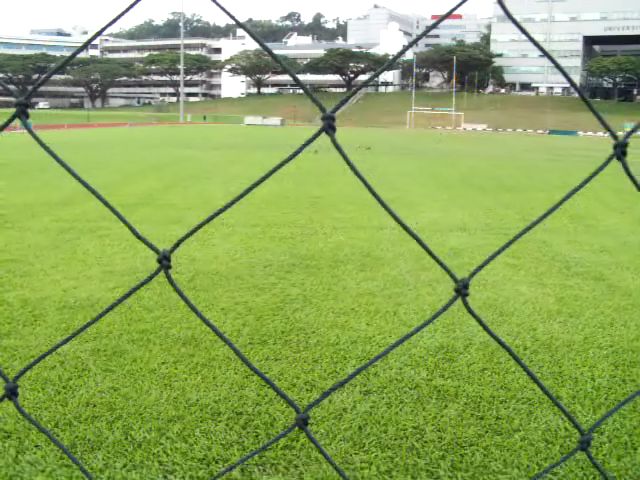}&
			\includegraphics[height=1.6cm]{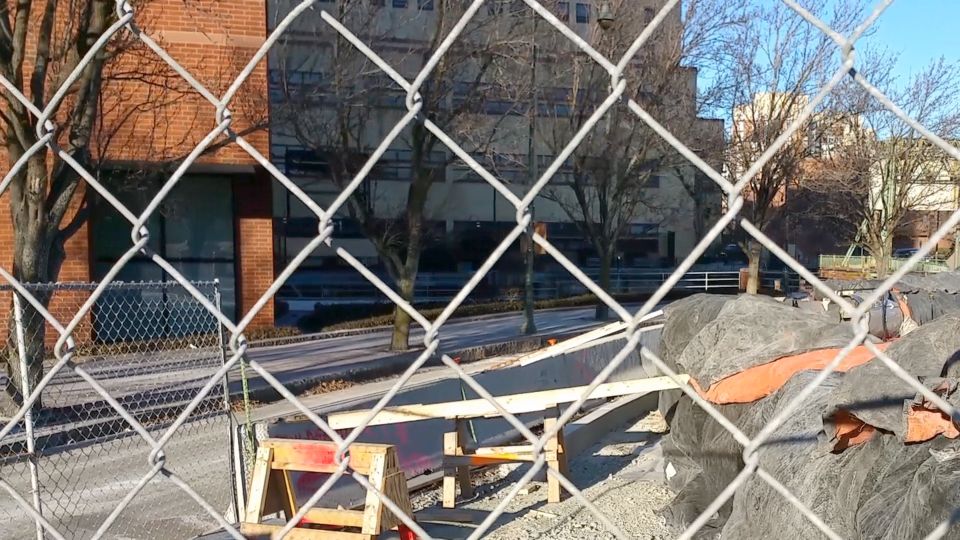}&
			\includegraphics[height=1.6cm]{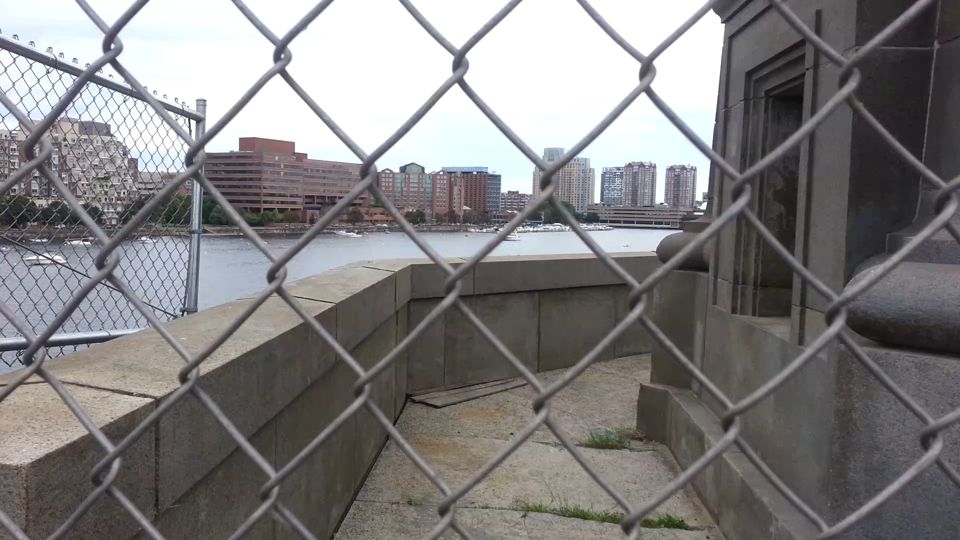}&
			\includegraphics[height=1.6cm]{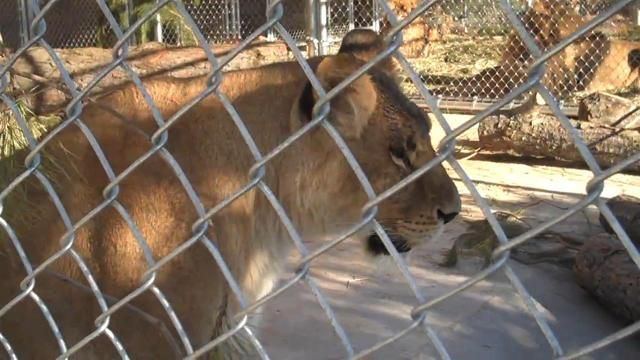}&
			\includegraphics[height=1.6cm]{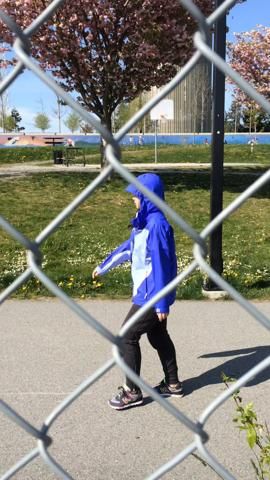}\\
			\includegraphics[height=1.6cm]{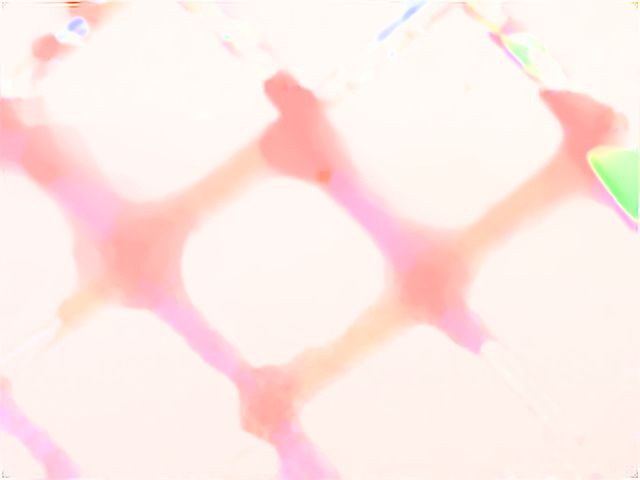}&
			\includegraphics[height=1.6cm]{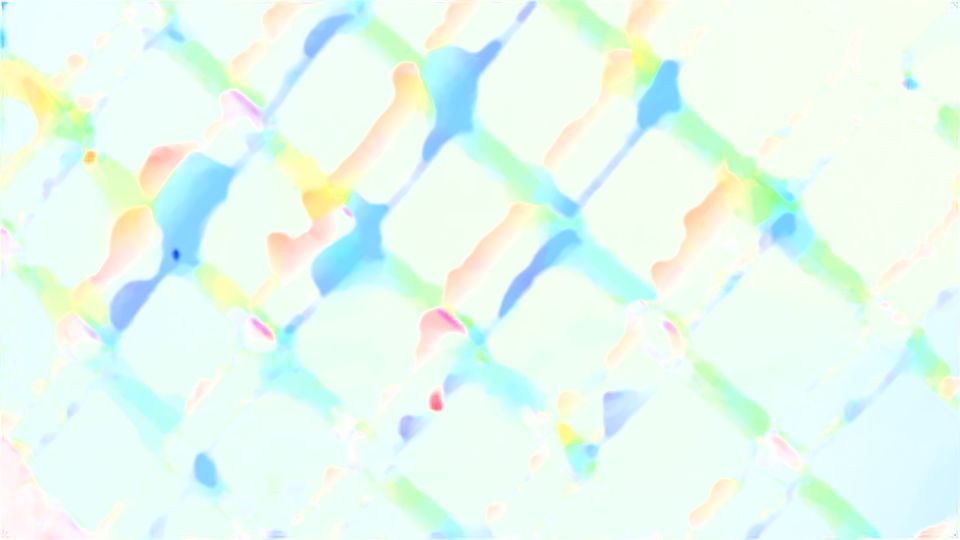}&
			\includegraphics[height=1.6cm]{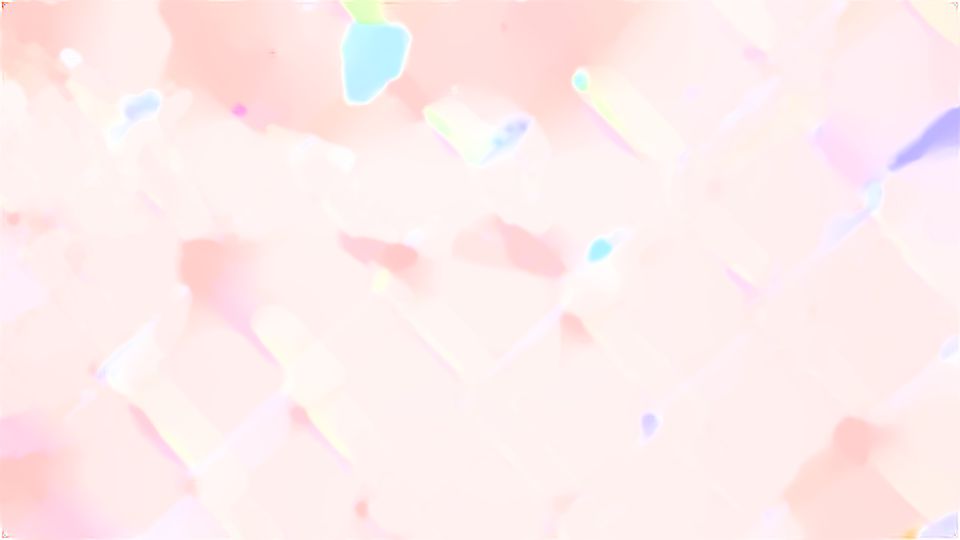}&
			\includegraphics[height=1.6cm]{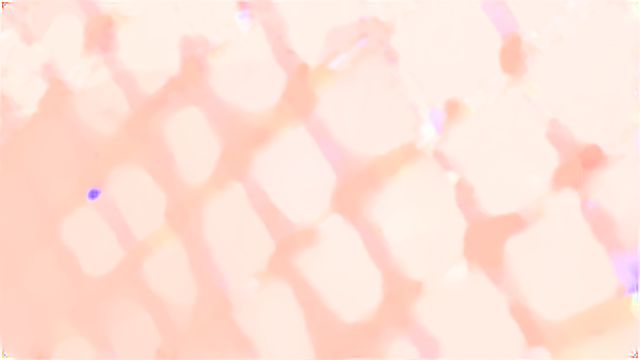}&
			\includegraphics[height=1.6cm]{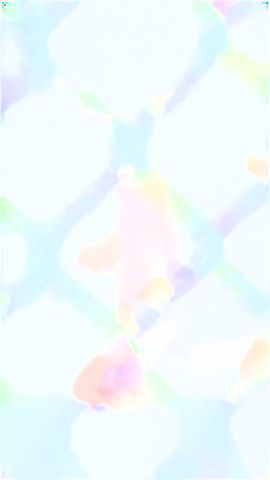}\\
			\includegraphics[height=1.6cm]{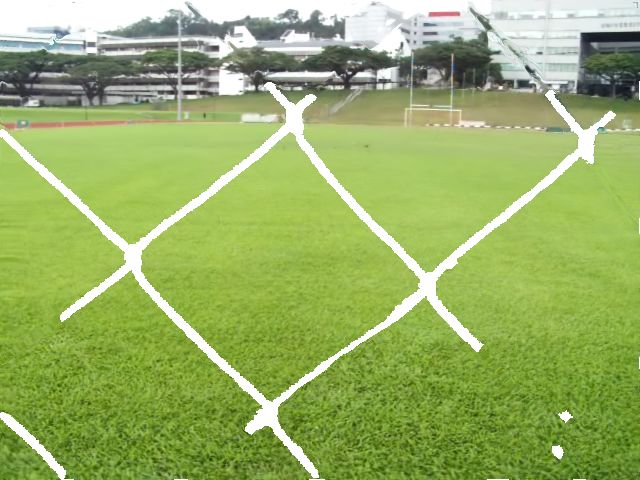}&
			\includegraphics[height=1.6cm]{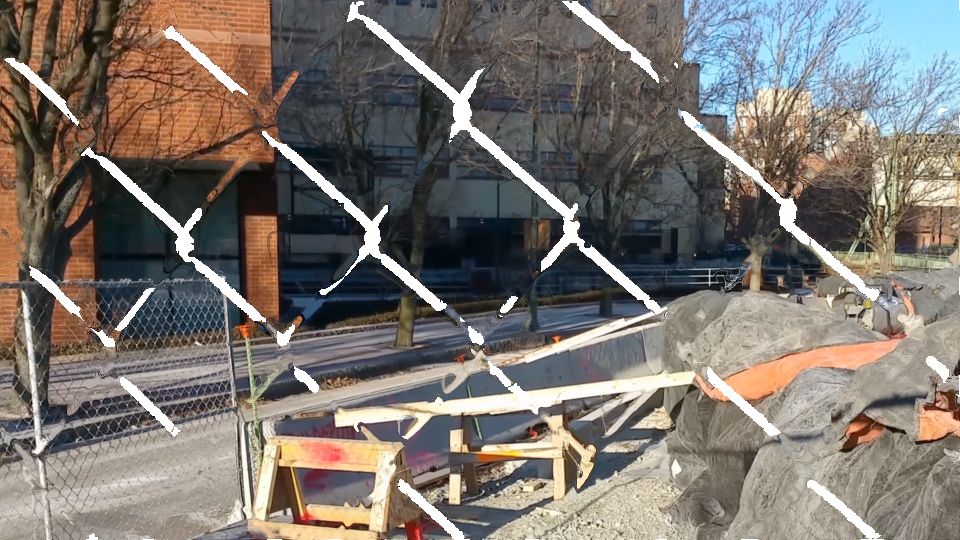}&
			\includegraphics[height=1.6cm]{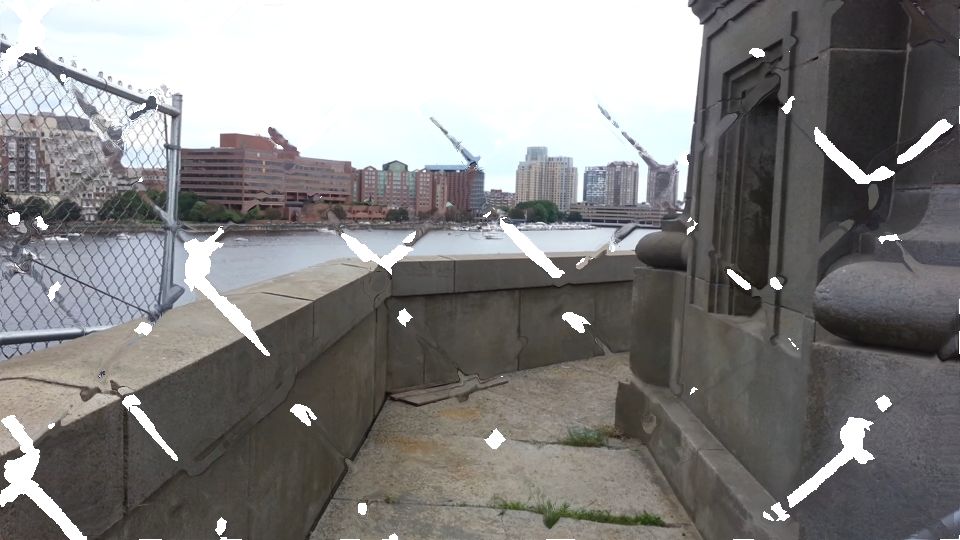}&
			\includegraphics[height=1.6cm]{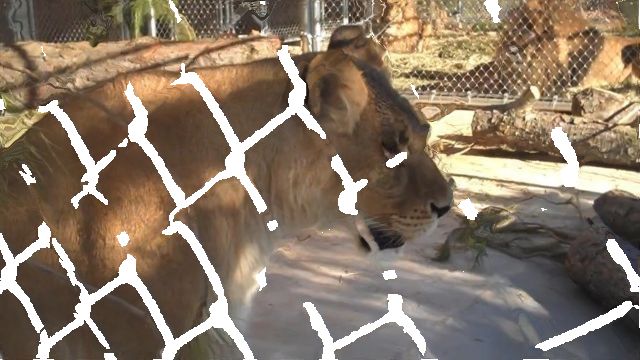}&
			\includegraphics[height=1.6cm]{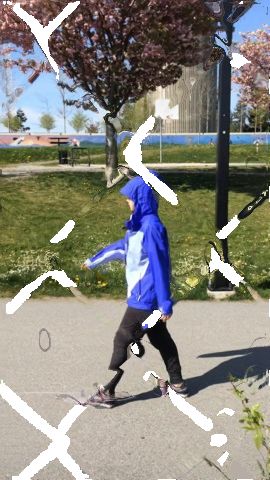}\\
			\includegraphics[height=1.6cm]{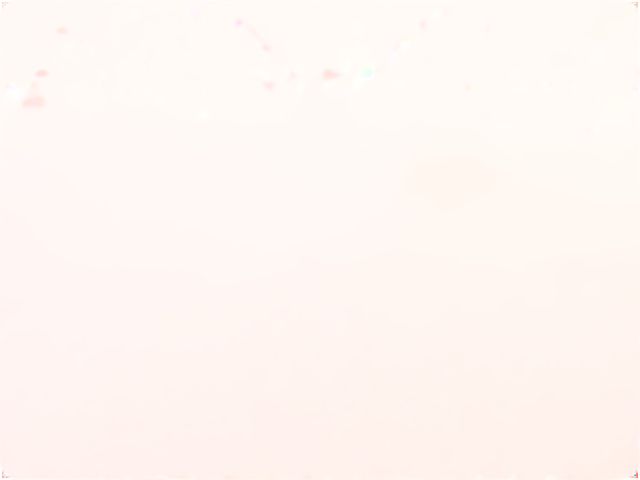}&
			\includegraphics[height=1.6cm]{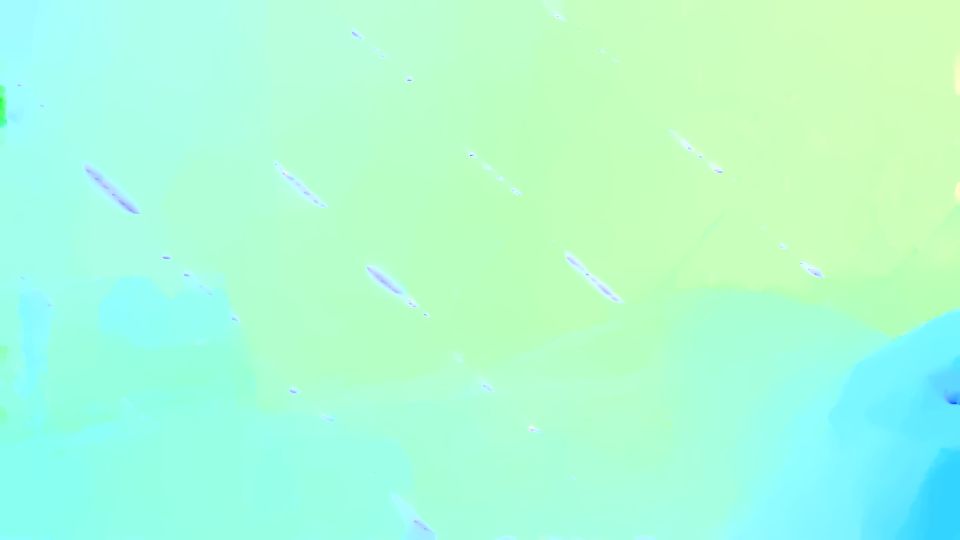}&
			\includegraphics[height=1.6cm]{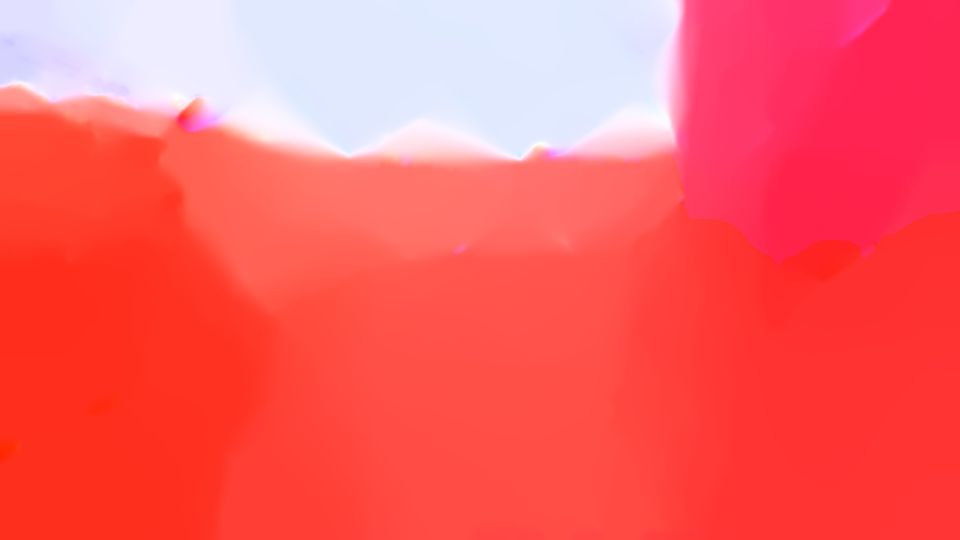}&
			\includegraphics[height=1.6cm]{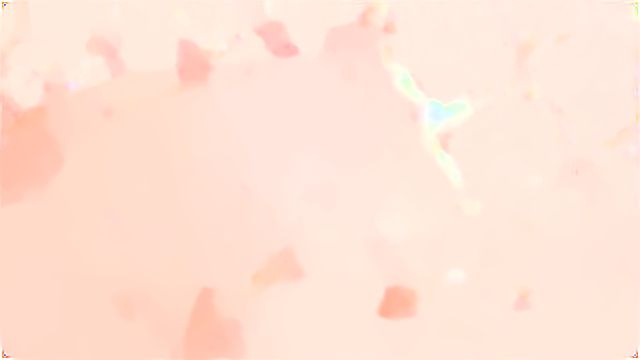}&
			\includegraphics[height=1.6cm]{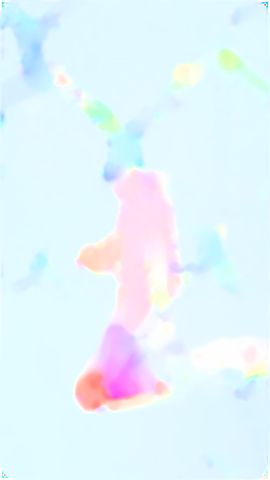}\\
		\end{tabular}
		\caption{ First and second row: frames taken from videos reported in \cite{CVPR_2016,Yadong,acm_sigg2015}. Third row: optical flow computed between the first and second row images using \cite{Brox}. Fourth row: de-fenced images obtained using the estimated flow shown in the third row. Fifth row: occlusion-aware optical flow obtained using the proposed algorithm.}
		\label{fig:flow}
	\end{figure}

	To demonstrate the robustness of proposed optical flow algorithm under known occlusions, we use frames from videos of fenced scenes in \cite{CVPR_2016,Yadong,acm_sigg2015}. We show two frames from a video sequence named \textquotedblleft football" from \cite{Yadong} in the first column of Fig. \ref{fig:flow}. The video sequences named \textquotedblleft fence1"  and \textquotedblleft fence4" are taken from the work of \cite{acm_sigg2015}. Two frames from each of these videos are shown in second and third columns of Fig. \ref{fig:flow}, respectively. Video sequences named \textquotedblleft lion" and \textquotedblleft walking" are taken from \cite{CVPR_2016} and a couple of observations from each of them are depicted in fourth and fifth columns of Fig. \ref{fig:flow}, respectively. In the third row of Fig. \ref{fig:flow}, we show the color coded optical flows obtained using \cite{Brox} between respective images shown in each column of first and second row of Fig. \ref{fig:flow}. Note that the images shown in third row of Fig. \ref{fig:flow} contain regions of erroneously estimated optical flow due to fence occlusions. In contrast, the flow estimated using proposed algorithm under known fence occlusions are shown in the fifth row of Fig. \ref{fig:flow}. Note that the optical flows estimated using the proposed method contain no artifacts.

	\subsection{Image De-fencing}
	
\begin{figure}[!htb]
		\centering
		\begin{tabular}{c c c}
			\includegraphics[width=3.5cm]{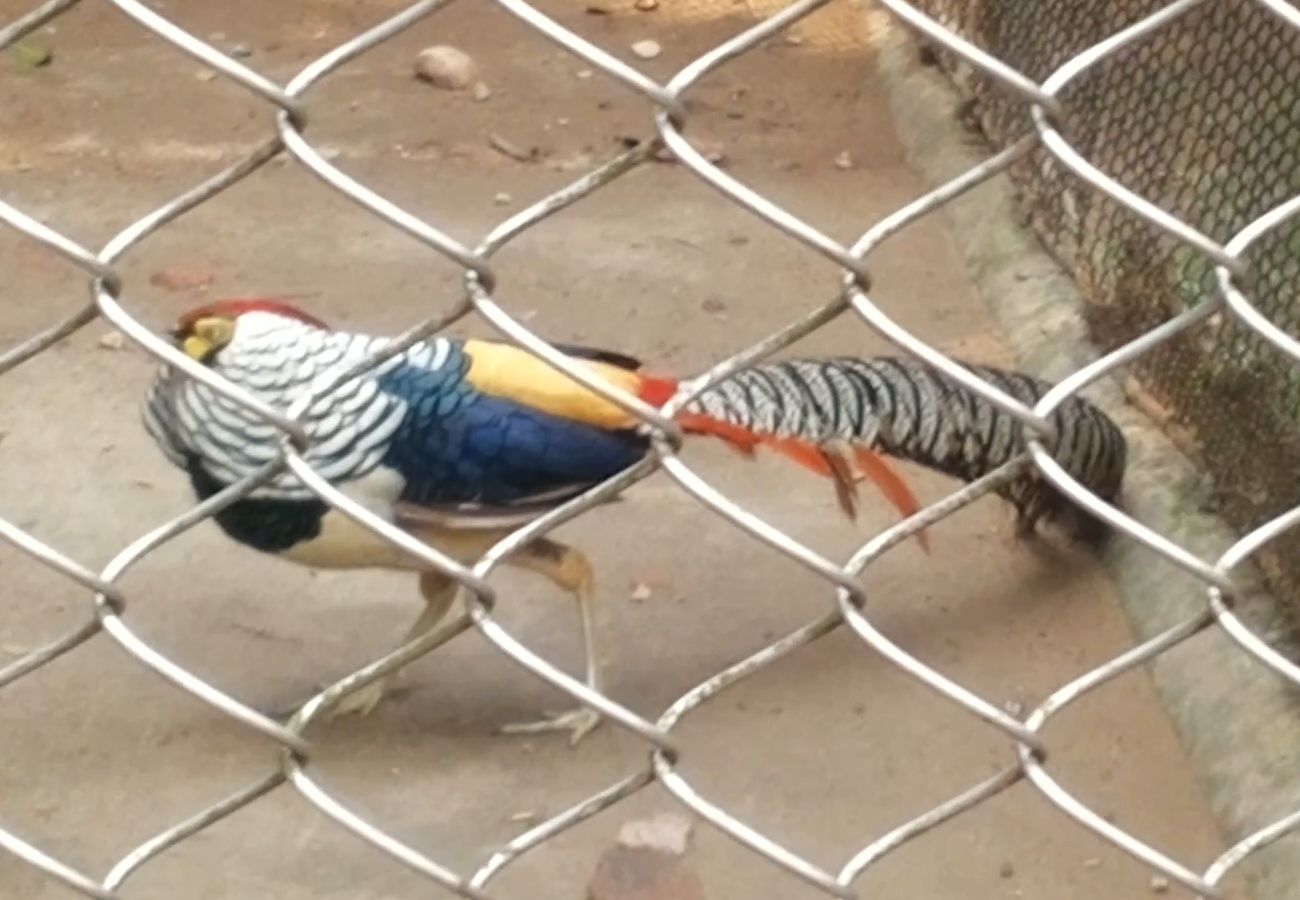}&
			\includegraphics[width=3.5cm]{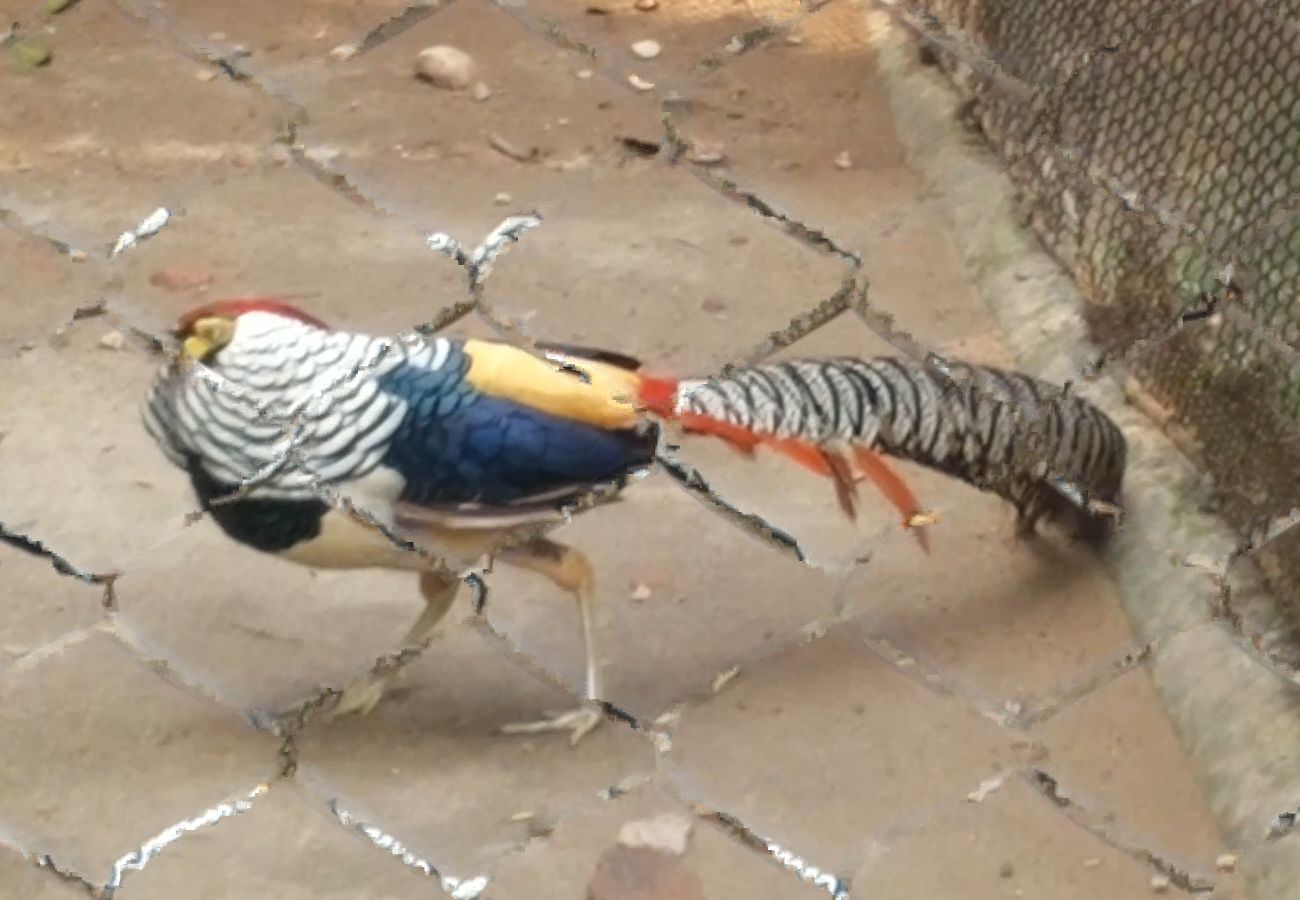}&
			\includegraphics[width=3.5cm]{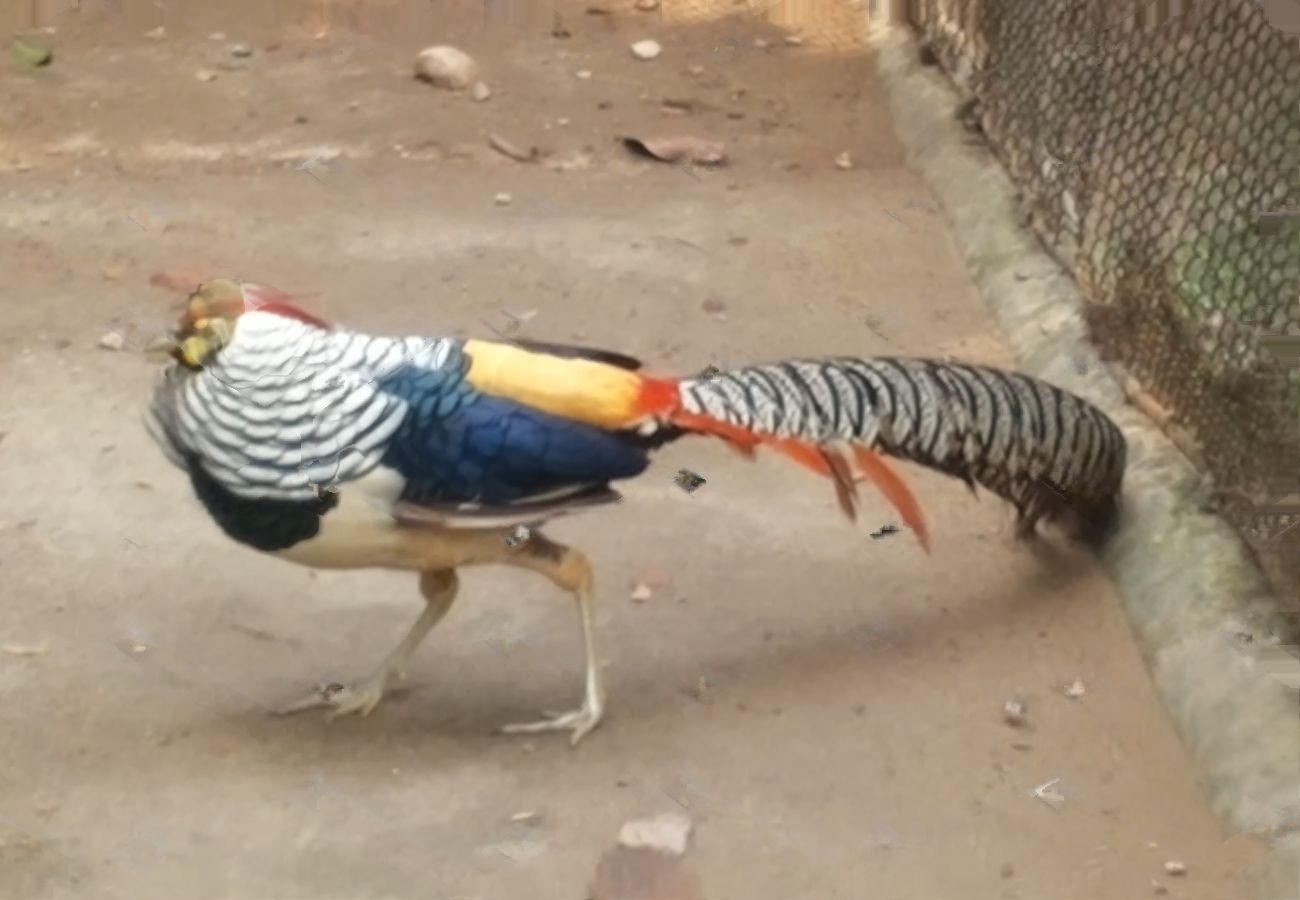}\\
			(a) & (b) & (c) \\							
			\includegraphics[width=3.5cm]{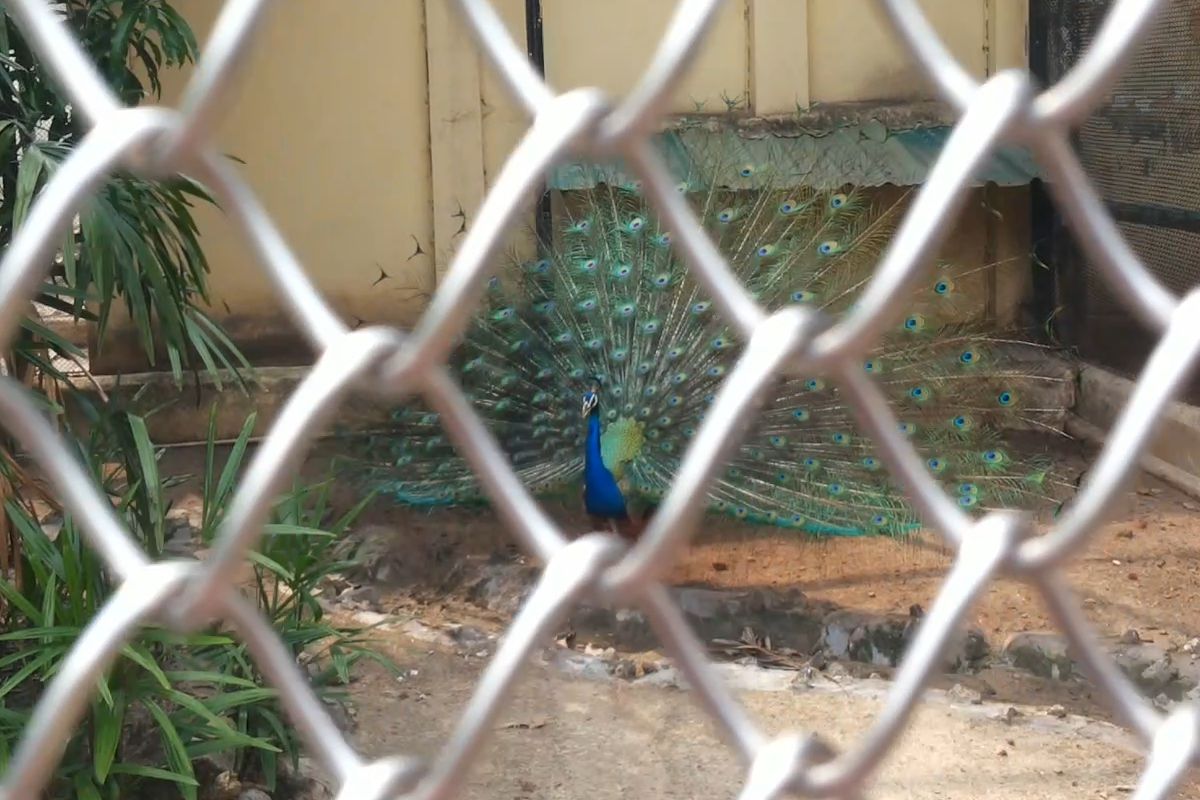}&
			\includegraphics[width=3.5cm]{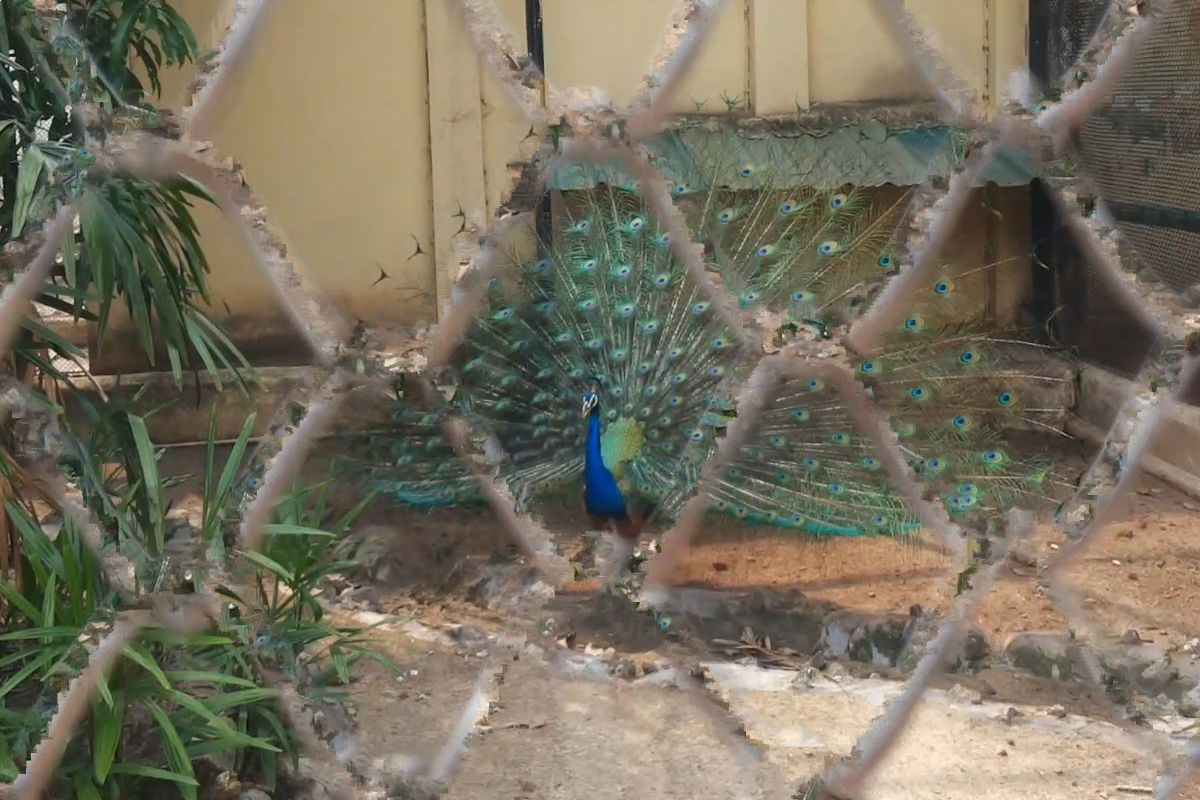}&
			\includegraphics[width=3.5cm]{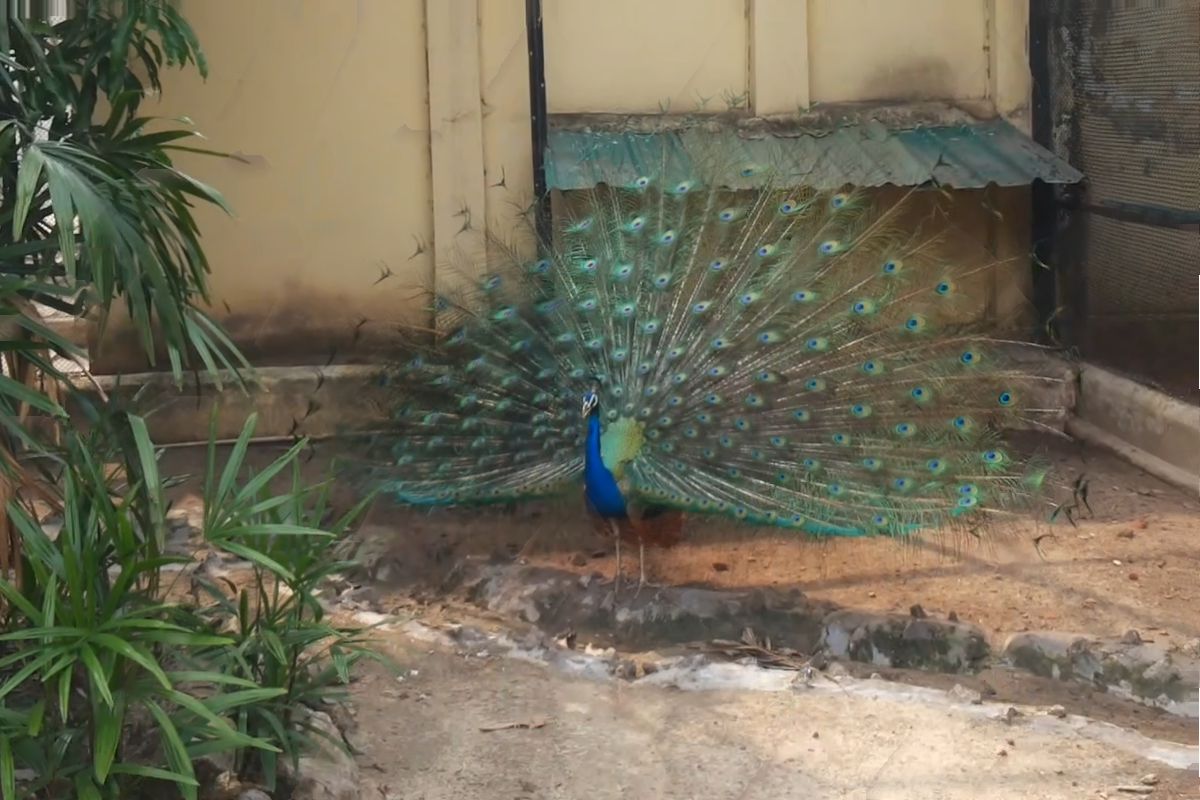}\\
			(d) & (e) & (f) \\								
			\includegraphics[width=3.5cm]{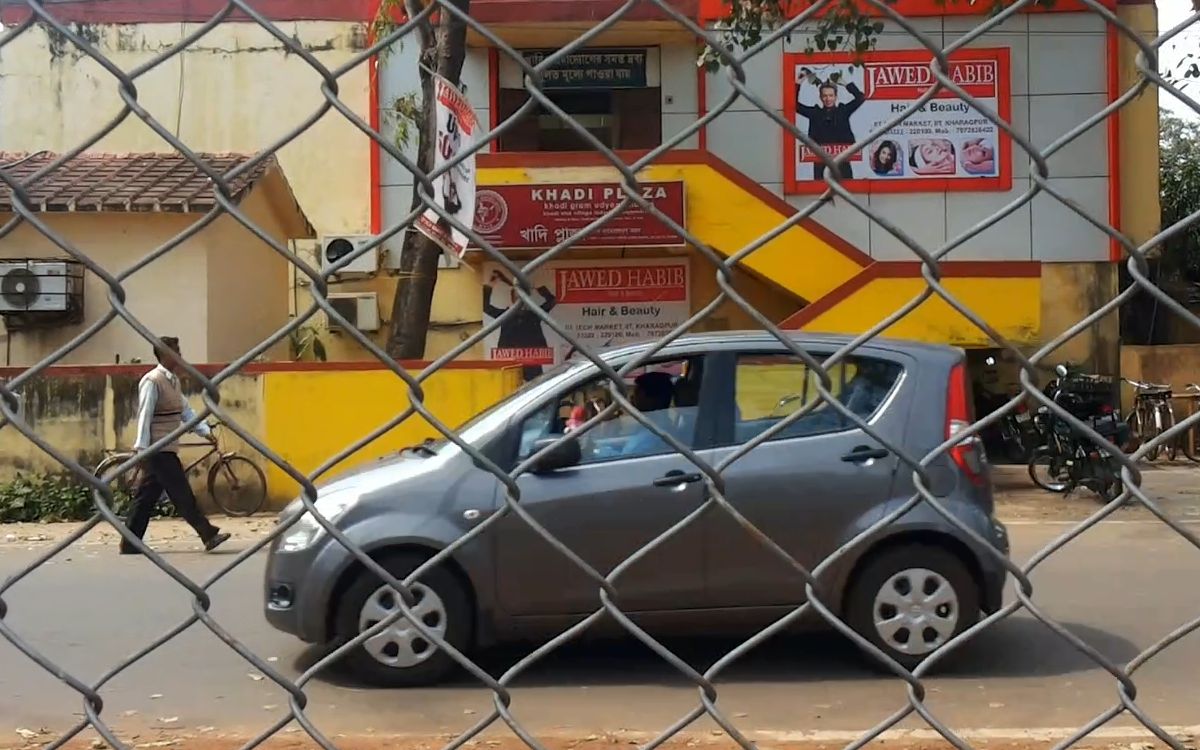}&
			\includegraphics[width=3.5cm]{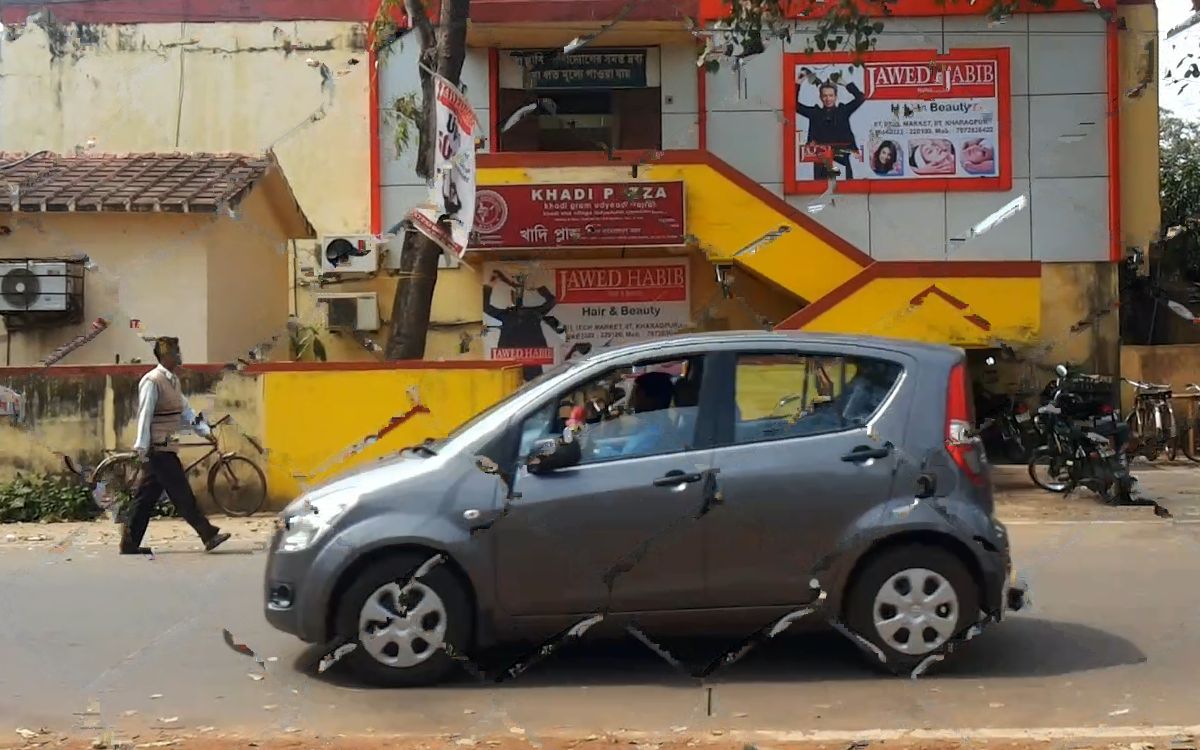}&
			\includegraphics[width=3.5cm]{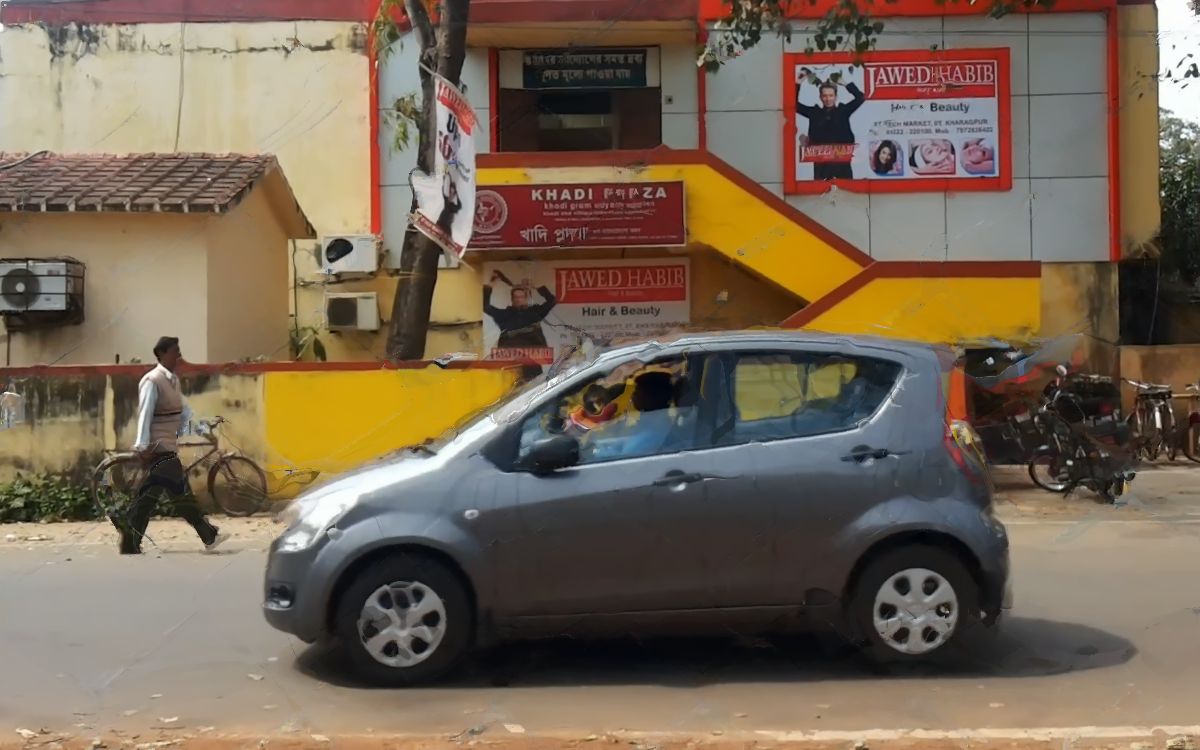}\\
			(g) & (h) & (i) \\	
			\includegraphics[width=3.5cm]{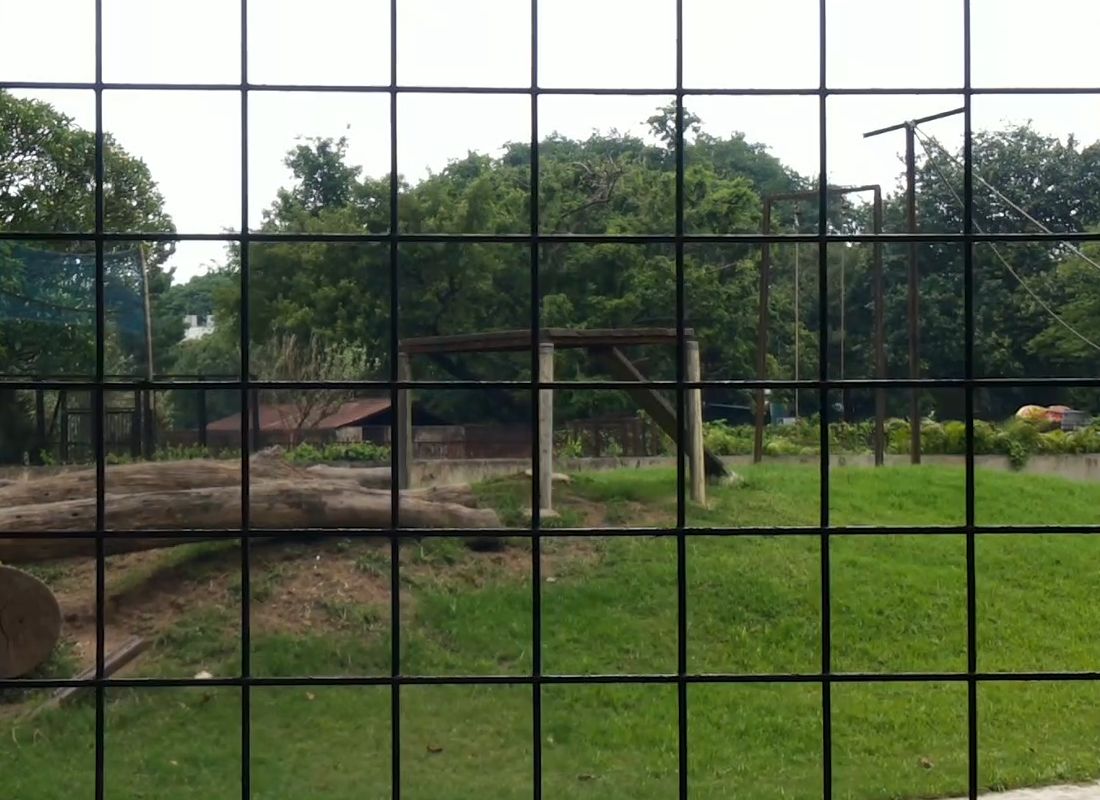}&
			\includegraphics[width=3.5cm]{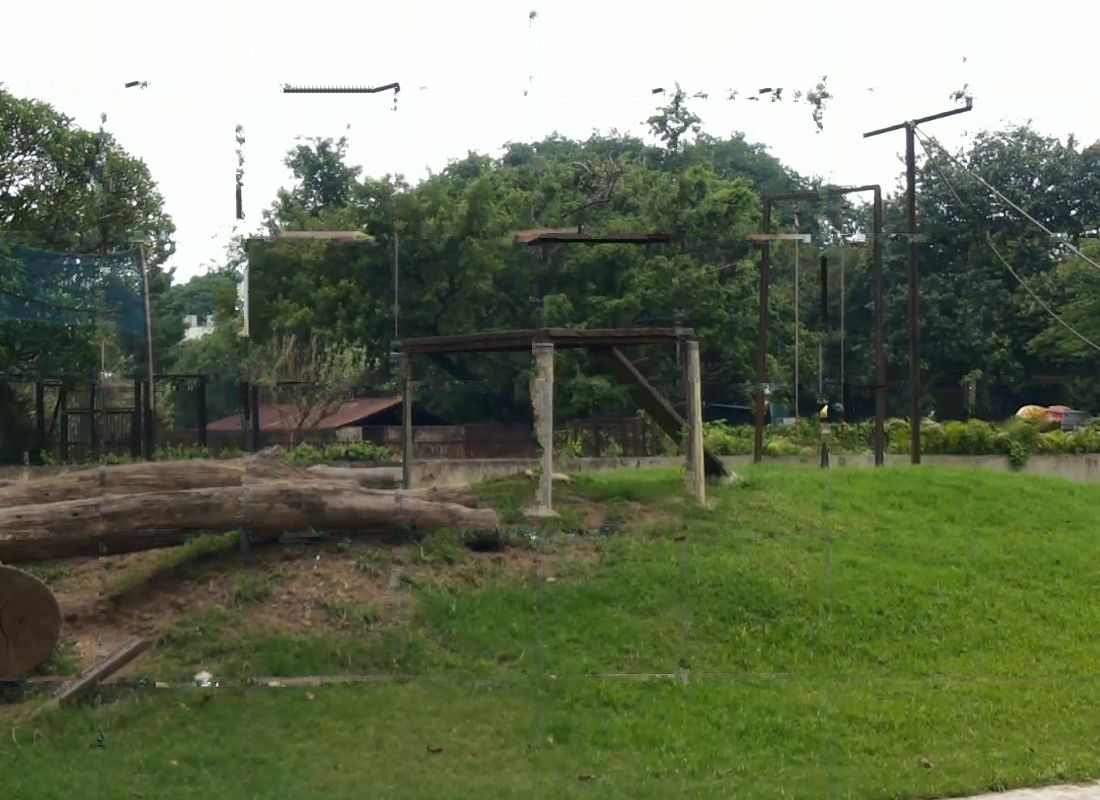}&
			\includegraphics[width=3.5cm]{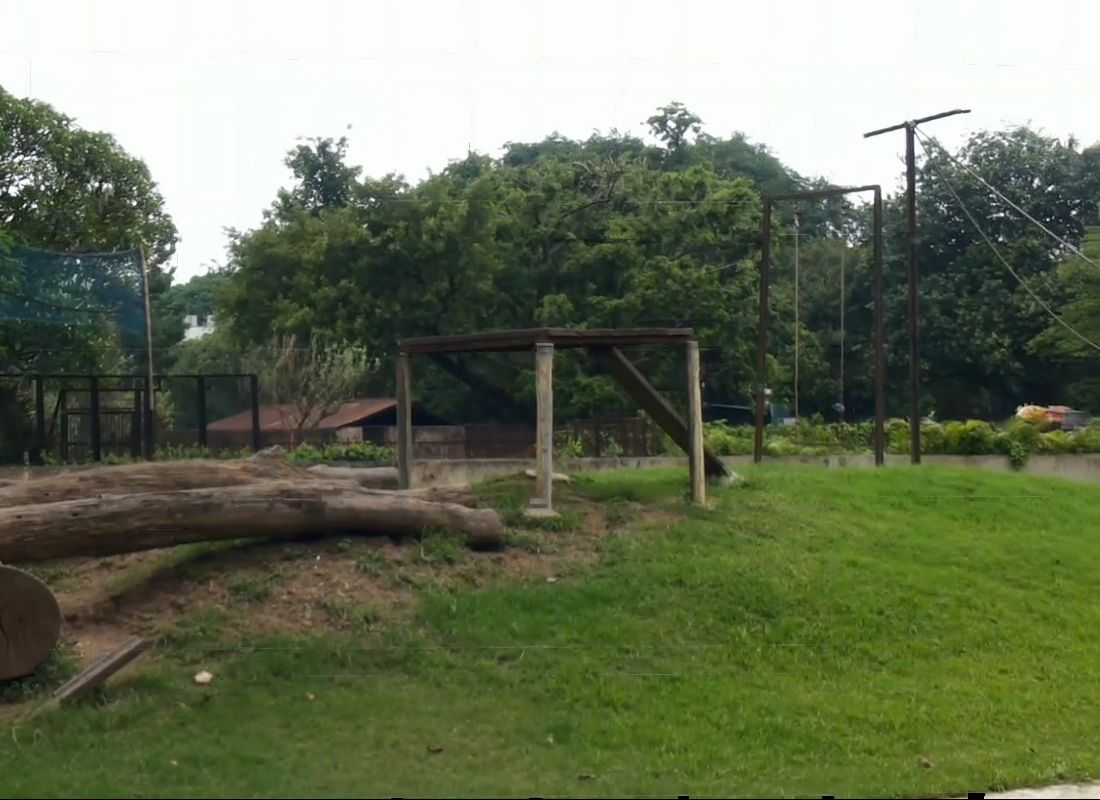}\\
			(j) & (k) & (l) \\								
		\end{tabular}
		\caption{ First column: one frame each taken from challenging real-world videos. Second column: inpainted images obtained using exemplar-based image inpainting algorithm \cite{Criminisi} which was the approach used in \cite{CVPR_2016} for image de-fencing. Third column: de-fenced images obtained using the proposed algorithm corresponding to images in the first column.}
		\label{fig:defencing}
	\end{figure}

	To demonstrate the efficacy of the proposed image de-fencing algorithm, we conducted experiments with several real-world video sequences containing dynamic background objects. In Figs. \ref{fig:defencing} (a), (d), (g), and (j), we show the images taken from four different video sequences. The fence pixels corresponding to these observations are segmented using the proposed pre-trained CNN-SVM and the approach of \cite{Yuanjie}. In Figs. \ref{fig:defencing} (b), (e), (h), and (k), we show the inpainted images obtained using \cite{Criminisi} which was the method used for obtaining the de-fenced image after fence segmentation in \cite{CVPR_2016}. Note that we can see several artifacts in the inpainted images obtained using \cite{Criminisi}. De-fenced images obtained using the proposed algorithm are shown in Figs. \ref{fig:defencing} (c), (f), (i), and (l), respectively. We observe that the proposed algorithm has effectively reconstructed image data even for dynamic real-world video sequences. Also, note that for all the results shown in Figs. \ref{fig:defencing} (c), (f), (i), and (l) we used only three observations from the captured video sequences. 	
	
\begin{figure}[!htb]
		\centering
		\begin{tabular}{c c c c}	
			\includegraphics[height=1.8cm]{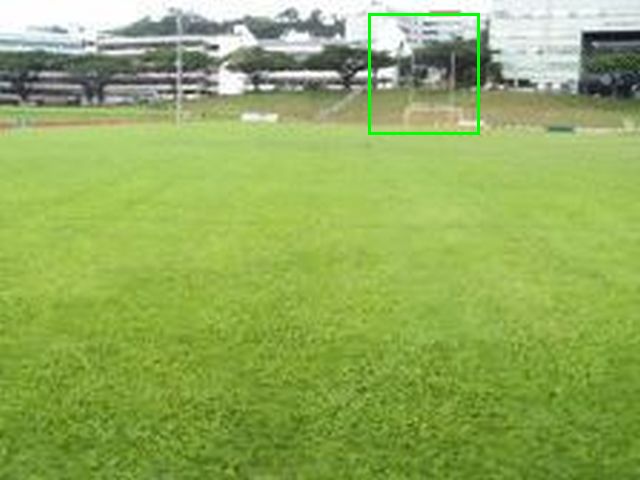}&
			\includegraphics[height=1.8cm]{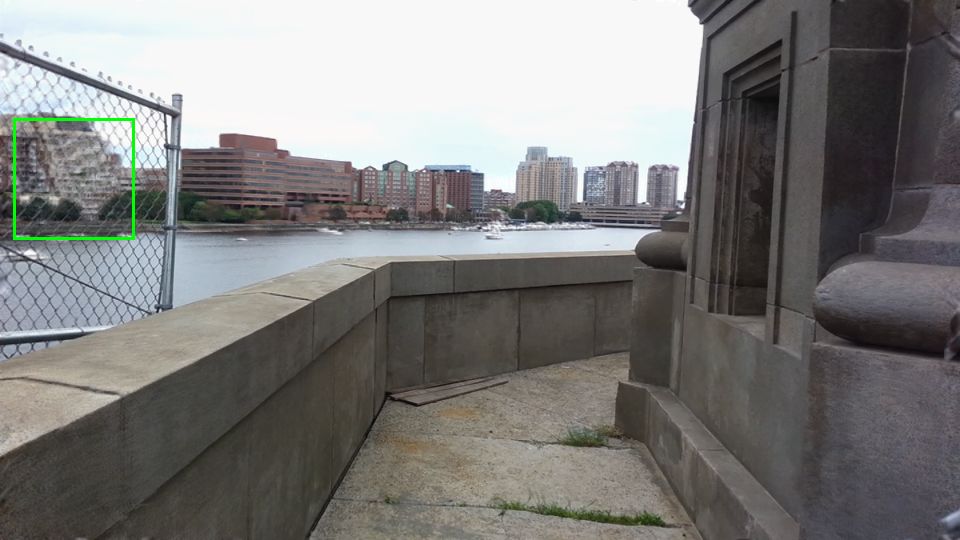}&
			\includegraphics[height=1.8cm]{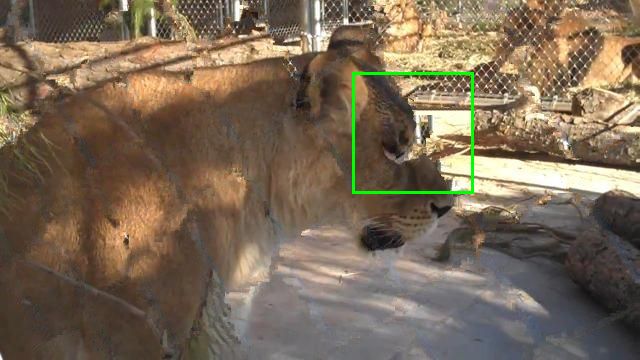}&
			\includegraphics[height=1.8cm]{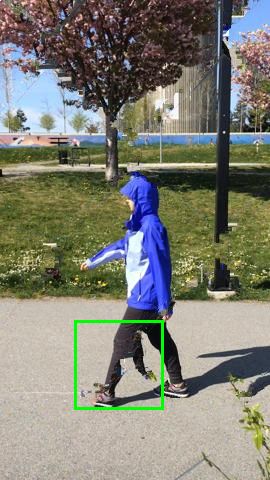}\\
			(a) & (b) & (c) & (d)\\	
			\includegraphics[height=1.8cm]{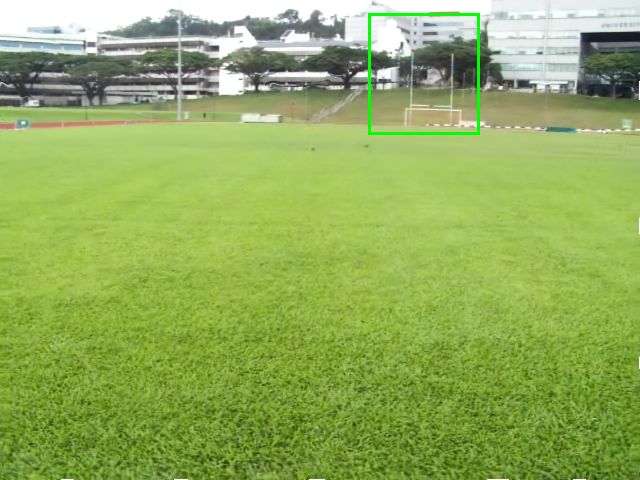}&
			\includegraphics[height=1.8cm]{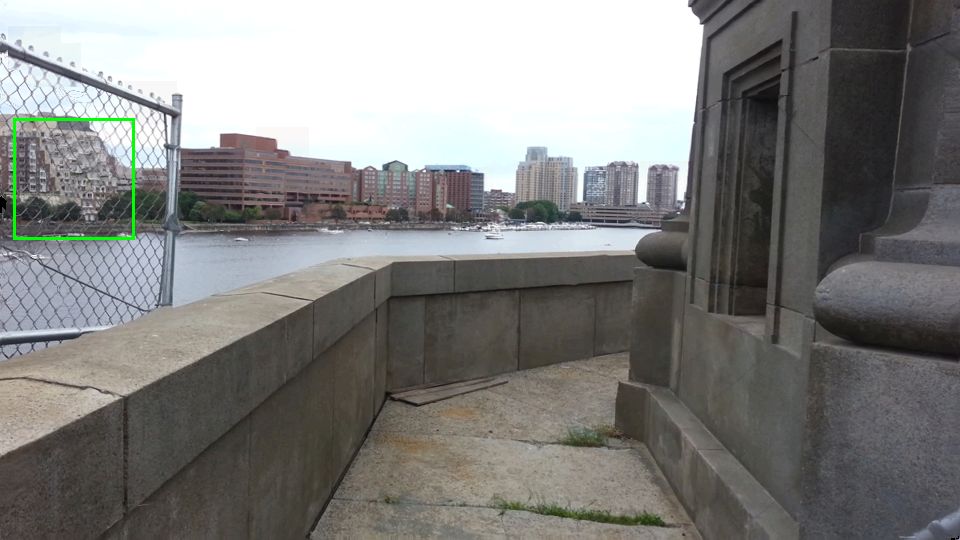}&
			\includegraphics[height=1.8cm]{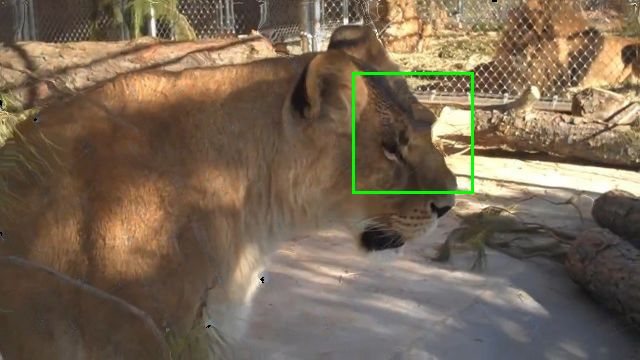}&
			\includegraphics[height=1.8cm]{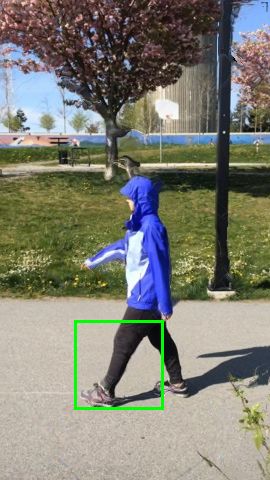}\\
			(e) & (f) & (g) & (h)\\	
		\end{tabular}
		\begin{tabular}{c c c c c c c c}	
			\includegraphics[height=1.1cm]{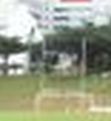}&
			\includegraphics[height=1.2cm]{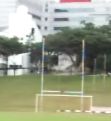}&
			\includegraphics[height=1.2cm]{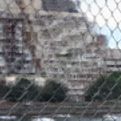}&
			\includegraphics[height=1.2cm]{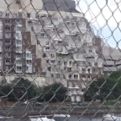}&
			\includegraphics[height=1.2cm]{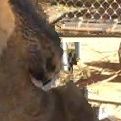}&
			\includegraphics[height=1.2cm]{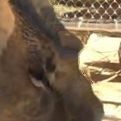}&
			\includegraphics[height=1.2cm]{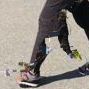}&	\includegraphics[height=1.2cm]{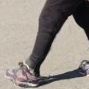}\\	
		\end{tabular}
		\caption{Comparison with state-of-the-art image/video de-fencing methods \cite{Yadong,acm_sigg2015,CVPR_2016} using video sequences from their works. (a) De-fenced image obtained by \cite{Yadong}. (b) Recovered background image using \cite{acm_sigg2015}. (c), (d) Inpainted images obtained by \cite{Criminisi} which was the method used in \cite{CVPR_2016}.  (e)-(h) De-fenced images obtained by the proposed algorithm using occlusion-aware-optical flow shown in fifth row of Fig. \ref{fig:flow}. Last row: Insets from the images of first and second rows, respectively, showing the superior reconstruction of the de-fenced image by our algorithm.}
		\label{fig:comp}
	\end{figure}

	Next, we compare the proposed algorithm with recent state-of-the-art methods \cite{CVPR_2016,Yadong,acm_sigg2015}. In Fig. \ref{fig:comp} (a), we show the de-fenced image obtained using \cite{Yadong}. The corresponding result obtained by the proposed algorithm is shown in Fig. \ref{fig:comp} (e). Note that the de-fenced image obtained in \cite{Yadong} is blurred whereas the proposed algorithm generated a sharper image. We show a cropped region from both Figs. \ref{fig:comp} (a) and (e) in the last row to confirm our observation. In Figs. \ref{fig:comp} (b) and (f), we show the de-fenced results obtained by \cite{acm_sigg2015} and the proposed algorithm, respectively. The de-fenced image obtained using the method in \cite{acm_sigg2015} is distorted at some places which is apparent in Fig. \ref{fig:comp} (b). In contrast, the fence has been removed completely with hardly any distortions in the result shown in Fig. \ref{fig:comp} (f), which  has been obtained using our algorithm. A cropped region from both Figs. \ref{fig:comp} (b) and (f) are shown in the last row to prove our point. The de-fenced images obtained using a very recent technique \cite{CVPR_2016} are shown in Figs. \ref{fig:comp} (c) and (d), respectively. These results contain several artifacts. However, the de-fenced images recovered using the proposed algorithm hardly contain any artifacts as shown in Figs. \ref{fig:comp} (g) and (h). A cropped regions from Figs. \ref{fig:comp} (c) and (d) and Figs. \ref{fig:comp} (g) and (h) are shown in the last row for comparison purpose. Since we use only three frames from the videos, our method is more computationally efficient than \cite{Yadong,acm_sigg2015} which use $5$ and $15$ frames, respectively.
	
	\section{Conclusions}
	
	In this paper, we proposed an automatic image de-fencing system for real-world videos.  We divided the problem of image de-fencing into three tasks and proposed an automatic approach for each one of them. We formulated an optimization framework and solved the inverse problem using the fast iterative shrinkage thresholding algorithm (FISTA) assuming $l_{1}$ norm of the de-fenced image as the regularization constraint. We have evaluated the proposed algorithm on various datasets and reported both qualitative and quantitative results. The obtained results show the effectiveness of proposed algorithm. As part of future work, we are investigating how to optimally choose the frames from the video for fence removal.

\clearpage

\bibliographystyle{splncs}
\bibliography{eccv_db}
\end{document}